%% file: iclr2027_conference.tex
\documentclass{article} 
\usepackage{iclr2027_conference,times}

\input{math_commands.tex}

\usepackage{float}
\usepackage{graphicx}
\usepackage{subcaption}

\usepackage{hyperref}
\usepackage{url}
\usepackage{booktabs}
\usepackage{multirow}
\usepackage{graphicx}
\usepackage{array}
\usepackage{subcaption}

\title{Beyond Anonymous Captions: Grounding Character Identity in Video Captioning and Question Answering}

\author{
\textbf{Anas Filali Razzouki}\textsuperscript{1,2}
\quad
\textbf{Killian Steunou}\textsuperscript{1,2}
\quad
\textbf{Khalil Guetari}\textsuperscript{2}
\quad
\textbf{Thomas Kling}\textsuperscript{2}
\\
\textbf{Mounîm El-Yacoubi}\textsuperscript{1}
\quad
\textbf{Yannis Tevissen}\textsuperscript{1,2}
\\[2mm]
\textsuperscript{1}Télécom SudParis, Institut Polytechnique de Paris, France\\
\textsuperscript{2}Moments Lab Research, France
}
\iclrfinalcopy 
\begin{document}

\maketitle

\begin{abstract}
Linking people's appearance and actions to character identities is essential for understanding video narratives. We present a framework for identity-aware video captioning and person-centric question answering that combines automatic character identification, explicit spatial grounding, and task-specific adaptation. Starting from LSMDC v2 movie clips, our pipeline matches detected faces to actor reference images, tracks characters across frames, and builds inputs with identity-linked bounding boxes. A strong vision-language model generates identity-aware captions and questions, which are manually verified and filtered to create a benchmark of 750 captioned clips and 3,000 person-centric questions. We study five grounding strategies combining textual coordinates with visual face or estimated person boxes across Video-MLLM families at roughly 2B, 4B, and 8B parameters and larger frontier models. Combining visual face boxes with textual coordinates yields the most consistent performance across scales and significantly improves overall performance over coordinates alone. Smaller models tend to over-assign known identities when the queried person is not grounded, while larger models better recognize such UNIDENTIFIED cases. We introduce BAC by LoRA fine-tuning Qwen models at 2B, 4B, and 8B scales on about 32K identity-aware captioned clips. Across all scales, BAC outperforms every other evaluated model family of comparable size. BAC-8B reaches 93.20\% overall QA accuracy, ranking behind only GPT-5.6 Sol among the frontier models evaluated in our study. Overall, explicitly communicating \emph{who is where}, together with lightweight task-specific adaptation, substantially improves identity-aware video understanding without changing the underlying architecture. We release the benchmark, training data, code, and BAC checkpoints at \url{https://github.com/momentslab/beyond-anonymous-captions}.
\end{abstract}

\section{Introduction}
Understanding human-centered video requires more than recognizing what happens in a scene. Video-captioning models can describe events, appearances, actions, and interactions in natural language~\citep{xu2016msr,Yang_2023_CVPR}. However, these descriptions are typically identity-agnostic: a model may correctly understand what each visible person is doing without determining which character that person corresponds to. In narrative videos such as movies, this distinction is important because scene understanding depends not only on recognizing actions and interactions, but also on attributing them to the correct characters.

At the same time, visual grounding has made substantial progress in associating textual references with corresponding visual regions~\citep{Kamath_2021_ICCV,ma2024groma}. Given a textual reference to a person or object, grounding models can localize the referred entity, typically through spatial regions such as bounding boxes. Video captioning therefore provides a mechanism for describing what is happening in a scene, while visual grounding provides complementary information about where a referred entity is located.

This observation motivates the central question of our work: can these two capabilities be combined to support identity-aware video understanding? More specifically, if a Video-MLLM is provided not only with the visual content of a video, but also with explicit information indicating the identity and spatial location of the people appearing in it, can it correctly associate each identity with the corresponding person in the scene? Such an association could allow the model to preserve its existing understanding of appearances, actions, spatial relations, and interactions while grounding that understanding in the correct character identities. More generally, we investigate whether explicit information about \emph{who is where} can help a Video-MLLM determine more reliably \emph{who is doing what, where, and with whom}.

Prior work on identity-aware visual description has approached the problem in several ways. Early movie-naming methods such as M-VAD Names~\citep{pini2019m} associate face tracks with character identities and replace generic \emph{SOMEONE} mentions in existing captions with the corresponding names, rather than generating identity-aware descriptions directly. A related post-processing strategy is proposed by Tevissen et al.~\citep{tevissen2024inserting}, who first generate a generic image caption and then use attention maps and identified face regions to replace person mentions with names. In both cases, identity is introduced after the caption content has largely been determined, so people omitted from the original caption cannot naturally be recovered. M-VAD Names indeed formulates its main task as replacing \emph{SOMEONE} tags in existing captions with proper character names.

More recent work integrates identity more directly into multimodal understanding. MICap~\citep{raajesh2024micap} maintains anonymous person identities across a videoset of five consecutive clips by clustering faces and jointly learning identity assignment and caption generation. IDA-VLM~\citep{ji2025ida}, instead, conditions a vision-language model on reference images of known identities and requires the model to recognize those identities in new scenes before performing tasks such as localization, question answering, or captioning. ISYV~\citep{gao2026seek} extends this reference-conditioned setting to video, requiring the model to recognize and reason about a target person across time and shot transitions. Thus, these methods either learn identity consistency jointly with generation or require the multimodal model itself to infer the correspondence between a reference identity and its occurrence in the visual input.

A complementary line of work studies region-level multimodal grounding. Omni-RGPT~\citep{heo2025omni} associates user-specified boxes or masks with language references to support region-specific reasoning over images and videos, while VideoGLaMM~\citep{munasinghe2025videoglamm} grounds language in video at the pixel level through spatio-temporally consistent segmentation masks. These approaches demonstrate that explicit links between language and visual regions can support fine-grained reasoning, but they are not designed specifically to determine how known character identities should be represented to a general-purpose Video-MLLM.

Despite these advances, existing approaches do not address our central question: how should already known character identities be spatially represented to a general-purpose Video-MLLM so that it can reason about them throughout a video? We introduce this formulation by providing explicit identity-linked spatial cues whenever reliable detections and tracks are available, rather than requiring the model to recover identity solely from reference images or appearance. Importantly, these cues may be absent in some frames, requiring the Video-MLLM to propagate the available identity information across time and associate each character with their appearances, actions, locations, and interactions. We therefore investigate how different spatial representations of known identities enable identity-aware video captioning and person-centric question answering, without designing a dedicated region-aware architecture.

To study this question, we develop a pipeline on movie shots from LSMDC v2 in which detected faces are matched to actor reference images and identified characters are spatially grounded across sampled frames. We construct a manually verified benchmark of 750 captioned shots and 3,000 person-centric questions, and compare five identity-grounding representations ranging from textual coordinates to visual face or person boxes and their combinations. We evaluate them on identity-aware video captioning and person-centric question answering, then use the resulting grounding formulation to introduce BAC at three model scales (2B, 4B, and 8B). We compare BAC with Video-MLLMs of similar sizes from multiple model families and extend the evaluation to frontier open-weight and proprietary video-language models. We release approximately 32K synthetically captioned training clips and a manually verified benchmark of 750 clips and 3,000 questions, all with face bounding-box annotations, plus source code and trained checkpoints.

\section{Methods}

\subsection{Dataset}

\subsubsection{Character Identification}
We build our dataset from the LSMDC v2 collection~\citep{lsmdc,lsmdc2016MovieAnnotationRetrieval,maharaj2017dataset}, downloading \textbf{92 movies} comprising \textbf{46,756 video clips}. Following cast-supervised character-identification approaches for movies and TV series~\citep{xu2010cast2face,nagrani2018benedict,bamman2024measuring}, we construct character-linked face tracks by combining cast information, actor reference images, face recognition, and temporal tracking.

For each movie, we retrieve actor and character names from IMDb and collect up to 50 reference images per actor from DuckDuckGo. We filter low-quality or ambiguous images using InsightFace~\citep{deng2019arcface} based on image resolution, face size, detection confidence, and face dominance. The remaining face embeddings are clustered to remove identity outliers, and the 5 most representative images are retained for each actor, forming  a reliable reference gallery. 

Each clip is then processed frame-by-frame with InsightFace to detect faces and extract 512-dimensional embeddings. A DeepSORT-inspired online tracker~\citep{wojke2017simple} links detections across consecutive frames into temporally consistent tracks. Each track is represented by a robust aggregate embedding and matched against the movie-specific reference gallery. The predicted identity is accepted when the matching confidence exceeds \textbf{0.30} and is propagated across the track together with its face bounding boxes.

Finally, because the original LSMDC clips may contain multiple camera shots with abrupt changes in scene or viewpoint, we split them into visually coherent shots using PySceneDetect. This produces \textbf{54,133 shots}, with an average duration of \textbf{2.50\,s},  and an average of \textbf{1.7 shots per clip}.

\subsubsection{Caption and Question Generation}
\label{subsec:Caption_and_Question Generation}

\begin{figure*}[h]
    \centering
    \includegraphics[width=\textwidth]{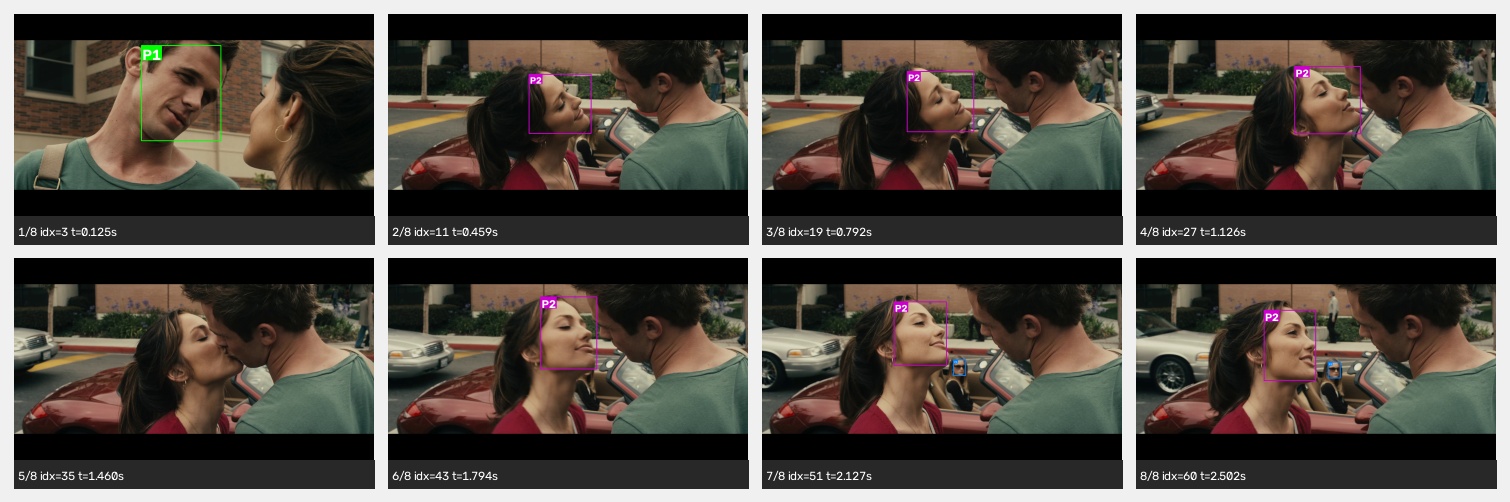}

    \vspace{1mm}
    \begin{minipage}{\textwidth}
    \small

\textbf{Identity mapping:}
\textcolor{green!60!black}{P1} $\leftrightarrow$ \textcolor{green!60!black}{Cam Gigandet};
\textcolor{magenta!80!black}{P2} $\leftrightarrow$ \textcolor{magenta!80!black}{Minka Kelly};
\textcolor{blue}{P3} $\leftrightarrow$ \textcolor{blue}{Leighton Meester}.

    \vspace{1mm}

    \textbf{Caption:}
    \textcolor{green!60!black}{Cam Gigandet}, in a green shirt, and
    \textcolor{magenta!80!black}{Minka Kelly}, in a red top, stand close together
    and share a kiss while \textcolor{blue}{Leighton Meester},
    in the car, watches.

    \vspace{1mm}

\textbf{Person-centric QA:}
\textit{Who is in the car?}
\textit{(Position)}
$\rightarrow$ \textcolor{blue}{Leighton Meester};
\textit{Who is kissing \textcolor{magenta!80!black}{Minka Kelly}?}
\textit{(Interaction)}
$\rightarrow$ \textcolor{green!60!black}{Cam Gigandet};
\textit{Who is wearing a green shirt?}
\textit{(Appearance)}
$\rightarrow$ \textcolor{green!60!black}{Cam Gigandet};
\textit{Who is wearing a red top?}
\textit{(Appearance)}
$\rightarrow$ \textcolor{magenta!80!black}{Minka Kelly}.

\end{minipage}

\caption{Example of identity-aware captioning and person-centric QA using character identities grounded across sampled video frames.}
    \label{fig:qualitative_qa}
\end{figure*}

Given the short shot duration, averaging approximately 2.5 s at 24 fps, we uniformly sample 8 frames per shot. As illustrated in Figure~\ref{fig:qualitative_qa}, these frames provide sufficient temporal coverage of the main actions. From the 54,133 detected shots, we retain only those containing at least one identity-linked bounding box in at least one sampled frame. Preserving the original LSMDC v2 movie-level split, this yields 31,935 shots from 72 training movies and 6,219 shots from 12 test movies. The training set is further divided into 28,742 training and 3,193 validation shots, with the validation split used for BAC hyperparameter tuning.

For both training and test shots, the sampled frames are provided to Gemini 2.5 Flash to generate identity-aware captions. For test shots, the model is additionally instructed to generate four person-centric questions and to assess shot validity, visual and temporal coherence, and identity-grounding difficulty. We retain shots that are valid, high-quality, and challenging, with coherence scores of at least 9 and challenge scores of at least 8.5. This favors non-trivial cases involving multiple people, distractors, occlusions, interactions, or viewpoint changes, yielding 2,008 shots.

Finally, from the 2,008 retained shots, we manually review and correct the captions, questions, and answers. We remove redundant or ambiguous examples and refine overly simple questions to require more precise identity grounding. This curation produces a final evaluation benchmark of 750 captioned shots and 3,000 person-centric questions.

\subsection{Identity-grounding representations}
\label{sec:grounding_representation}

Once character identities and their locations have been established, we investigate how these identity--location associations should be represented to a general-purpose Video-MLLM. We consider five grounding strategies that differ in whether spatial information is conveyed through textual coordinates, visual annotations, or a combination of both. Each identified character is assigned an identifier $P_i$, which is consistently mapped to the same identity throughout the shot.

\textbf{Face Textual Coordinates (FTC).}
The sampled frames are provided without visual bounding-box annotations. The prompt specifies the identity associated with each $P_i$ together with the textual coordinates of its face bounding box for every frame in which the character is localized. This representation evaluates whether the Video-MLLM can establish the identity--location association using textual spatial information alone.

For strategies involving visual grounding, a bounding box is drawn around each localized character, with the corresponding $P_i$ displayed at its upper-left corner. For a given identity, the box and label use the same color across frames, providing a consistent identity-linked visual cue. The $P_i$ label size is scaled with the bounding box to remain visible while limiting occlusion. Figure~\ref{fig:qualitative_qa} illustrates this visual identity-grounding process.

\textbf{Face Visual Boxes (FVB).}
Each identified character is visually grounded using its face bounding box. No bounding-box coordinates are provided in the textual prompt, such that the identity--location association is conveyed exclusively through the visual face annotations.

\textbf{Person Visual Boxes (PVB).}
Each identified character is visually grounded using an estimated person-level bounding box. Rather than requiring an additional person detector, we extrapolate the person box from the detected face using a lightweight anthropometric heuristic based on fixed face-to-body proportions~\citep{jaruenpunyasak2022anthropometric}. Details of the estimation procedure are provided in Appendix~\ref{app:pbbox}. In this method, no spatial coordinates are included in the prompt; identity--location associations are conveyed solely through person-level visual annotations.

\textbf{Face Visual Boxes + Face Textual Coordinates (FVB+FTC).}
Each identified character is visually grounded using its face bounding box. In addition to the visual annotation, the prompt provides the textual coordinates of the corresponding face bounding box for each frame.

\textbf{Person Visual Boxes + Person Textual Coordinates (PVB+PTC).}
Each identified character is visually grounded using an estimated person-level bounding box. In addition to the visual annotation, the prompt provides the textual coordinates of the corresponding estimated person bounding box for each frame.

Together, these five representations allow us to compare textual grounding, visual grounding, and their combination, while also examining whether face-level or person-level spatial cues are more effective for identity-aware video understanding.\subsection{Experimental Protocol}
\label{sec:experimental_protocol}

We evaluate the proposed identity-grounding representations on two complementary tasks: identity-aware video captioning and person-centric question answering. All experiments are conducted on the manually verified evaluation set of 750 video clips, comprising 3,000 identity-related questions.

We conduct three complementary sets of experiments. First, we study the effect of the grounding representation while keeping the Video-MLLM family fixed. We compare the five representations introduced above: FTC, FVB, PVB, FVB+FCT, and PVB+PTC. This comparison is performed with Qwen3 models at approximately 2B, 4B, and 8B parameters, allowing us to examine how the effectiveness of each grounding representation evolves with model scale.

Second, we investigate whether the benefits of identity grounding generalize across different Video-MLLM families. Models are grouped into three approximate parameter scales: 2B, 4B, and 8B--9B. At the 2B scale, we evaluate Qwen3, InternVL3.5; at the 4B scale, we evaluate Qwen3, InternVL3.5, MiniCPM-V, and Ovis2.5; and at the 8B--9B scale, we evaluate Qwen3, Ovis2, InternVL3.5, Eagle2.5, and Molmo2. The same clips, questions, prompts, identity annotations, and grounding representations are used across models within each comparison.

Third, we introduce BAC, obtained by fine-tuning the Qwen model family at three scales (2B, 4B, and 8B) using Low-Rank Adaptation (LoRA). The models are trained on our training split using $\approx 32\text{K}$ identity-aware captions generated by Gemini as supervision, as described in Section~\ref{subsec:Caption_and_Question Generation}, and are evaluated on the  manually verified evaluation set. This stage assesses whether task-specific, parameter-efficient adaptation can further improve identity-aware video understanding when combined with explicit identity grounding.

For captioning, each model receives the grounded video shot and is instructed to generate a description referring to identified characters using their corresponding $P_i$ identifiers. For question answering, the model receives the same grounded shot together with one person-centric question at a time and returns the corresponding $P_i$, or \texttt{UNIDENTIFIED (UNID)} when the queried person cannot be associated with any provided identity. The predicted $P_i$ identifiers are subsequently mapped to their corresponding character names for evaluation and qualitative presentation.

Except for the LoRA adaptation experiments, all models are evaluated without task-specific fine-tuning. We use a consistent prompting protocol within each task and keep the generation configuration fixed for each model across the compared grounding representations.
\subsection{Evaluation Metrics}
\label{sec:evaluation_metrics}
\textbf{Person-centric question answering:}
QA predictions are compared directly with the ground-truth identity. We report QA Overall Accuracy over all 3,000 questions, of which 2,074 target a known character and 926 correspond to UNIDENTIFIED cases. QA Named Accuracy is computed over the 2,074 questions targeting a known character, while QA UNIDENTIFIED Accuracy is computed over the 926 questions referring to a person who cannot be associated with any of the provided character identities. We additionally report QA Clip-Level Accuracy, defined as the percentage of clips for which all four associated questions are answered correctly.

\textbf{Identity-aware captioning:}
Evaluating identity accuracy against a single ground-truth caption is unreliable because two correct captions may describe different visible attributes of the same character, such as a shirt, skirt, or hat. Manual evaluation is also impractical at scale. We therefore use a VLM-as-a-judge protocol based on Qwen/Qwen3.8-27B-FP8, an open-weight 27B vision-language model achieving 94\% accuracy on our person-centric QA task. The judge receives the video shot, identity grounding, and generated caption, and evaluates identity correctness, hallucination, visual faithfulness, and naturalness. We report Caption Overall, the mean quality score on a 0--10 scale; Caption Identity, the percentage of shots where every queried character is mentioned and correctly identified; and Caption Hallucination, the percentage containing an unsupported person, identity, or absent object. 
\section{Results}
\subsection{Effect of Identity Grounding Strategies}
Table~\ref{tab:grounding_by_size} shows that explicit visual grounding substantially improves person-centric QA over FTC alone. QA Overall increases from 53.87\% to 68.90\% at 2B, from 62.57\% to 79.73\% at 4B, and from 59.77\% to 88.53\% at 8B. Paired McNemar tests confirm that FTC is significantly worse than every visually grounded variant at all model sizes ($p<0.001$). Excluding FTC, differences among the grounded variants become smaller with scale. At 2B, FVB+FTC achieves the highest accuracy and significantly outperforms PVB+PTC and PVB, while remaining statistically indistinguishable from FVB. At 4B and 8B, FVB+FTC, FVB, and PVB+PTC consistently form the top statistical group, while PVB remains significantly weaker.

For captioning, the effect of the grounding representation is most pronounced at small model scale. At 2B, FVB+FTC clearly outperforms the other strategies in identity accuracy, with all pairwise differences being statistically significant ($p<0.001$). As model size increases, these differences shrink substantially: at 4B, the visually grounded variants are mostly statistically indistinguishable, with only FVB+FTC outperforming FVB, while at 8B all four visually grounded strategies are statistically tied. In contrast, FTC remains significantly worse than every visually grounded variant at 4B and 8B ($p<0.001$).

Overall, FVB+FTC provides the most consistent captioning and QA performance across scales, particularly at 2B. We therefore analyze whether its remaining errors depend on grounded face size. For QA, face-box area correlates positively with correctness, increasing from \(r_s=0.103\) at 2B to \(r_s=0.245\) at 8B, while captioning shows a stronger and stable association (\(r_s=0.238\)--\(0.256\)). PVB+PTC is less sensitive to box size, likely because the full-person box covers a much larger visual region than the face alone.

Results grouped by size confirm this pattern. At 2B (Table~\ref{tab:face_size_2b}), PVB+PTC performs better for small faces (\(<4{,}000\) px$^2$), reaching 91.8\% versus 80.0\% for FVB+FTC (\(p=0.021\)). For medium and large faces, FVB+FTC performs better, reaching 92.8\% vs.\ 91.3\% and 96.8\% vs.\ 91.1\%, respectively. These results suggest that full-person grounding can compensate when the face is very small, whereas face-based grounding becomes more effective once the face is sufficiently visible; expanding the grounding to the full person may then introduce less precise spatial information. Qualitative examples of both cases are provided in Appendix~\ref{app:effect_size}.

The Named and UNIDENTIFIED results further reveal how grounding affects identity assignment. At 2B and 4B, the combined visual--textual strategies favor assigning known identities: FVB+FTC and PVB+PTC achieve the highest Named accuracy, while remaining substantially weaker on UNIDENTIFIED cases. Visual-only grounding generally reduces this imbalance, improving the rejection of unidentified characters at the cost of lower Named accuracy. This asymmetry decreases markedly with model scale. At 8B, FVB+FTC achieves nearly balanced performance between Named and UNIDENTIFIED identities (88.04\% and 86.93\%), indicating that larger models are better able to use the grounding signal without systematically forcing a known identity when the evidence is insufficient.

\begin{table*}[h]
\centering
\footnotesize
\setlength{\tabcolsep}{4pt}
\renewcommand{\arraystretch}{0.95}

\caption{
Comparison of identity-grounding strategies across Qwen3-VL model sizes.
Captioning is evaluated using overall quality, identity accuracy, and
hallucination rate. QA performance is reported over all questions,
separately for Named and  UNIDENTIFIED (UNID) identities.
}
\label{tab:grounding_by_size}

\resizebox{0.85\textwidth}{!}{%
\begin{tabular}{llccc|ccc}
\toprule
& &
\multicolumn{3}{c|}{\textbf{Captioning}} &
\multicolumn{3}{c}{\textbf{Person-Centric QA}} \\
\cmidrule(lr){3-5}
\cmidrule(lr){6-8}

\textbf{Size} &
\textbf{Grounding} &
\textbf{Overall} $\uparrow$ &
\textbf{Identity} $\uparrow$ &
\textbf{Halluc.} $\downarrow$ &
\textbf{Overall} $\uparrow$ &
\textbf{Named} $\uparrow$ &
\textbf{UNID} $\uparrow$ \\
\midrule

\multirow{5}{*}{\textbf{2B}}
& FTC      & 5.74 & 47.47 & 7.60  & 53.87 & 75.60 & 5.18 \\
& FVB      & 5.92 & 36.80 & 10.27 & 68.27 & 86.55 & 27.32 \\
& PVB      & 5.52 & 27.54 & 8.28  & 66.13 & 80.71 & \textbf{33.48} \\
& FVB+FTC  & \textbf{7.61} & \textbf{76.53} & 3.07
           & \textbf{68.90} & \textbf{92.48} & 16.09 \\
& PVB+PTC  & 7.29 & 69.87 & \textbf{2.67}
           & 67.13 & 91.27 & 13.07 \\
\midrule

\multirow{5}{*}{\textbf{4B}}
& FTC      & 6.32 & 52.80 & 4.53 & 62.57 & 71.75 & 42.01 \\
& FVB      & 8.03 & 85.20 & 9.73 & 79.13 & 92.29 & 49.68 \\
& PVB      & 8.19 & 87.33 & 4.40 & 77.83 & 91.22 & 47.84 \\
& FVB+FTC  & \textbf{8.32} & \textbf{88.80} & \textbf{1.07}
           & \textbf{79.73} & 93.11 & \textbf{49.78} \\
& PVB+PTC  & 8.22 & 87.07 & 1.60
           & 79.20 & \textbf{94.21} & 45.57 \\
\midrule

\multirow{5}{*}{\textbf{8B}}
& FTC      & 6.32 & 54.27 & 3.60 & 59.77 & 51.69 & 77.86 \\
& FVB      & \textbf{8.45} & \textbf{92.00} & 3.87
           & \textbf{88.53} & 91.85 & 81.10 \\
& PVB      & 8.44 & 91.73 & 1.87 & 85.37 & 90.60 & 73.65 \\
& FVB+FTC  & \textbf{8.45} & 91.60 & \textbf{0.93}
           & 87.70 & 88.04 & \textbf{86.93} \\
& PVB+PTC  & 8.37 & 90.67 & 1.60
           & 87.43 & \textbf{92.24} & 76.67 \\
\bottomrule
\end{tabular}%
}
\end{table*}

\begin{table}[h]
\centering
\caption{
QA accuracy of FVB+FTC and PVB+PTC grounding methods with Qwen3-2B, stratified by face bounding-box area.
$\Delta$ denotes the accuracy difference between PVB+PTC and FVB+FTC.
}
\label{tab:face_size_2b}
\resizebox{0.85\textwidth}{!}{
\begin{tabular}{lrrrrr}
\toprule
\textbf{Face-area bin} &
\textbf{N} &
\textbf{FVB+FTC} &
\textbf{PVB+PTC} &
\textbf{$\Delta$ (P--F)} &
\textbf{$p$} \\
\midrule

\textbf{Small} $<4{,}000$ px$^2$
& 85
& 80.0\%
& \textbf{91.8\%}
& +11.8
& 0.021 \\

\textbf{Medium} $4{,}000$--$269{,}999$ px$^2$
& 1,865
& \textbf{92.8\%}
& 91.3\%
& -1.5
& 0.009 \\

\textbf{Large} $\geq270{,}000$ px$^2$
& 124
& \textbf{96.8\%}
& 91.1\%
& -5.6
& 0.016 \\

\bottomrule
\end{tabular}}
\end{table}

\definecolor{p4orange}{RGB}{230,126,34}

\definecolor{p4orange}{RGB}{230,126,34}

\definecolor{p4orange}{RGB}{230,126,34}

\begin{figure*}
    \centering

    \includegraphics[width=0.98\textwidth]
    {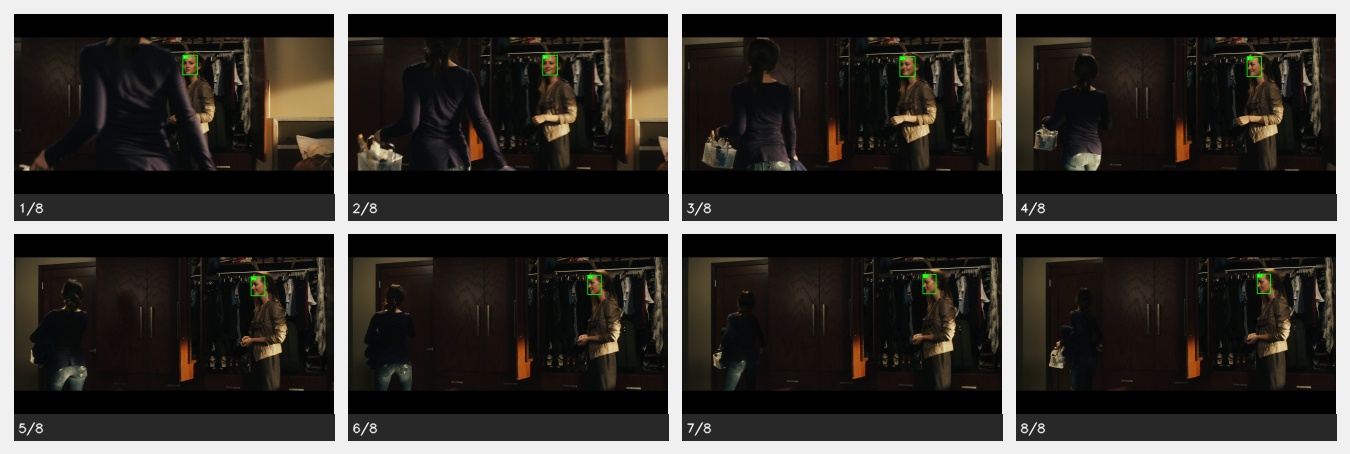}

    \vspace{0.3mm}

    \begin{minipage}{0.98\textwidth}
    \footnotesize

    \textbf{(a) Identity mapping:}
    \textcolor{green!60!black}{P1} $\leftrightarrow$
    \textcolor{green!60!black}{Leighton Meester}.

    \vspace{0.2mm}

    \textbf{Q4 [Action]:}
    \textit{Who walks away toward the door?}
    \hspace{0.6em}
    \textbf{GT:} \texttt{UNIDENTIFIED}.

    \vspace{0.3mm}

    \centering
    \scriptsize
    \setlength{\tabcolsep}{2.5pt}
    \renewcommand{\arraystretch}{0.90}

    \begin{tabular}{
        @{}lcc@{\hspace{8pt}}
        lcc@{\hspace{8pt}}
        lcc@{}
    }
        \toprule
        \textbf{Model} & \textbf{Pred.} & &
        \textbf{Model} & \textbf{Pred.} & &
        \textbf{Model} & \textbf{Pred.} & \\
        \midrule

        \textbf{BAC-2B (ours)} &
        \texttt{UNID.} &
        \textcolor{green!60!black}{$\checkmark$} &

        \textbf{BAC-4B (ours)} &
        \texttt{UNID.} &
        \textcolor{green!60!black}{$\checkmark$} &

        \textbf{BAC-8B (ours)} &
        \texttt{UNID.} &
        \textcolor{green!60!black}{$\checkmark$} \\

        Qwen3-VL 2B &
        Leighton &
        \textcolor{red}{$\times$} &

        Qwen3-VL 4B &
        Leighton &
        \textcolor{red}{$\times$} &

        Qwen3-VL 8B &
        Leighton &
        \textcolor{red}{$\times$} \\

        \midrule

        GPT-5.6 Sol &
        \texttt{UNID.} &
        \textcolor{green!60!black}{$\checkmark$} &

        GPT-5.6 Sol$^{*}$ &
        \texttt{UNID.} &
        \textcolor{green!60!black}{$\checkmark$} &

        Claude Sonnet 4.6 &
        \texttt{UNID.} &
        \textcolor{green!60!black}{$\checkmark$} \\

        Qwen3-VL 235B &
        \texttt{UNID.} &
        \textcolor{green!60!black}{$\checkmark$} &

        Gemini 2.5 Flash &
        \texttt{UNID.} &
        \textcolor{green!60!black}{$\checkmark$} &

        & & \\

        \bottomrule
    \end{tabular}

    \end{minipage}

    \vspace{1.2mm}

    \includegraphics[width=0.98\textwidth]
    {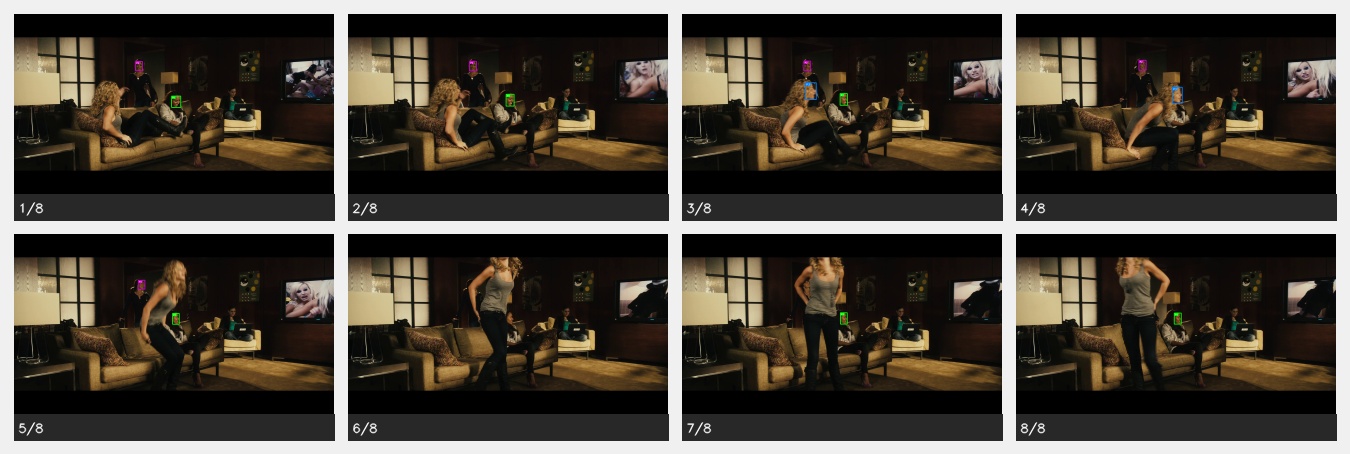}

    \vspace{0.3mm}

    \begin{minipage}{0.98\textwidth}
    \footnotesize

    \textbf{(b) Identity mapping:}
    \textcolor{green!60!black}{P1} $\leftrightarrow$
    \textcolor{green!60!black}{Kat Graham};
    \textcolor{magenta!80!black}{P2} $\leftrightarrow$
    \textcolor{magenta!80!black}{Minka Kelly};
    \textcolor{blue}{P3} $\leftrightarrow$
    \textcolor{blue}{Aly Michalka}.

    \vspace{0.2mm}

    \textbf{Q3 [Action]:}
    \textit{Who stands up from the brown sofa?}
    \hspace{0.6em}
    \textbf{GT:}
    \textcolor{blue}{Aly Michalka}
    (\textcolor{blue}{P3}).

    \vspace{0.3mm}

    \centering
    \scriptsize
    \setlength{\tabcolsep}{2.5pt}
    \renewcommand{\arraystretch}{0.90}

    \begin{tabular}{
        @{}lcc@{\hspace{8pt}}
        lcc@{\hspace{8pt}}
        lcc@{}
    }
        \toprule
        \textbf{Model} & \textbf{Pred.} & &
        \textbf{Model} & \textbf{Pred.} & &
        \textbf{Model} & \textbf{Pred.} & \\
        \midrule

        \textbf{BAC-2B (ours)} &
        Aly &
        \textcolor{green!60!black}{$\checkmark$} &

        \textbf{BAC-4B (ours)} &
        Aly &
        \textcolor{green!60!black}{$\checkmark$} &

        \textbf{BAC-8B (ours)} &
        Aly &
        \textcolor{green!60!black}{$\checkmark$} \\

        Qwen3-VL 2B &
        Kat &
        \textcolor{red}{$\times$} &

        Qwen3-VL 4B &
        Kat &
        \textcolor{red}{$\times$} &

        Qwen3-VL 8B &
        Minka &
        \textcolor{red}{$\times$} \\

        \midrule

        GPT-5.6 Sol &
        Aly &
        \textcolor{green!60!black}{$\checkmark$} &

        GPT-5.6 Sol$^{*}$ &
        Aly &
        \textcolor{green!60!black}{$\checkmark$} &

        Claude Sonnet 4.6 &
        \texttt{UNID.} &
        \textcolor{red}{$\times$} \\

        Qwen3-VL 235B &
        \texttt{UNID.} &
        \textcolor{red}{$\times$} &

        Gemini 2.5 Flash &
        \texttt{UNID.} &
        \textcolor{red}{$\times$} &

        & & \\

        \bottomrule
    \end{tabular}

    \end{minipage}

    \vspace{1.2mm}

    \includegraphics[width=0.98\textwidth]
    {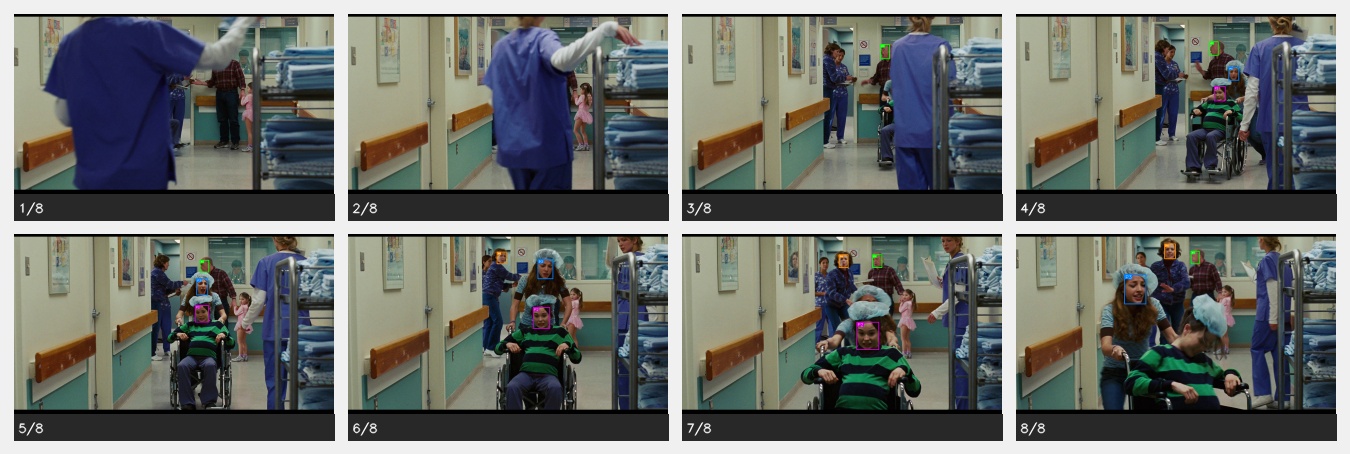}

    \vspace{0.3mm}

    \begin{minipage}{0.98\textwidth}
    \footnotesize

    \textbf{(c) Identity mapping:}
    \textcolor{green!60!black}{P1} $\leftrightarrow$
    \textcolor{green!60!black}{J. K. Simmons};
    \textcolor{magenta!80!black}{P2} $\leftrightarrow$
    \textcolor{magenta!80!black}{Elliot Page};
    \textcolor{blue}{P3} $\leftrightarrow$
    \textcolor{blue}{Olivia Thirlby};
    \textcolor{p4orange}{P4} $\leftrightarrow$
    \textcolor{p4orange}{Allison Janney}.

    \vspace{0.2mm}

    \textbf{Q4 [Position]:}
    \textit{Who follows behind Olivia Thirlby and Elliot Page?}
    \hspace{0.6em}
    \textbf{GT:}
    \textcolor{p4orange}{Allison Janney}
    (\textcolor{p4orange}{P4}).

    \vspace{0.3mm}

    \centering
    \scriptsize
    \setlength{\tabcolsep}{2.5pt}
    \renewcommand{\arraystretch}{0.90}

    \begin{tabular}{
        @{}lcc@{\hspace{8pt}}
        lcc@{\hspace{8pt}}
        lcc@{}
    }
        \toprule
        \textbf{Model} & \textbf{Pred.} & &
        \textbf{Model} & \textbf{Pred.} & &
        \textbf{Model} & \textbf{Pred.} & \\
        \midrule

        \textbf{BAC-2B (ours)} &
        J. K. &
        \textcolor{red}{$\times$} &

        \textbf{BAC-4B (ours)} &
        J. K. &
        \textcolor{red}{$\times$} &

        \textbf{BAC-8B (ours)} &
        J. K. &
        \textcolor{red}{$\times$} \\

        Qwen3-VL 2B &
        Elliot &
        \textcolor{red}{$\times$} &

        Qwen3-VL 4B &
        J. K. &
        \textcolor{red}{$\times$} &

        Qwen3-VL 8B &
        \texttt{UNID.} &
        \textcolor{red}{$\times$} \\

        \midrule

        GPT-5.6 Sol &
        Allison &
        \textcolor{green!60!black}{$\checkmark$} &

        GPT-5.6 Sol$^{*}$ &
        Allison &
        \textcolor{green!60!black}{$\checkmark$} &

        Claude Sonnet 4.6 &
        J. K. &
        \textcolor{red}{$\times$} \\

        Qwen3-VL 235B &
        \texttt{UNID.} &
        \textcolor{red}{$\times$} &

        Gemini 2.5 Flash &
        \texttt{UNID.} &
        \textcolor{red}{$\times$} &

        & & \\

        \bottomrule
    \end{tabular}

    \end{minipage}

    \vspace{-1.5mm}

    \caption{
    Qualitative identity-aware QA examples. Identity-mapping colors match the bounding-box colors in each frame.
    \textbf{(a)} BAC correctly predicts \texttt{UNIDENTIFIED};
    \textbf{(b)} BAC correctly identifies the queried named person;
    \textbf{(c)} BAC fails to identify the queried named person, while only GPT-5.6 Sol succeeds.
    $^{*}$: GPT-5.6 Sol with thinking.
    }

    \label{fig:bac_qualitative_examples}

    \vspace{-2mm}
\end{figure*}

\subsection{Benchmarking Across Model Families}

Table~\ref{tab:model_benchmark} compares BAC with several VLM families at comparable parameter scales using the FVB+FTC grounding method. We first focus on the base models to identify the strongest backbone before examining the effect of BAC fine-tuning. Among the base models, Qwen3 provides the strongest overall balance, achieving the highest QA accuracy within each size group together with consistently strong captioning performance. Its QA advantage over the strongest competing base model is statistically significant at 2B, 4B, and 8--9B (\(p<0.001\), paired McNemar tests). At 8--9B, Ovis2.5 attains higher caption identity accuracy than Qwen3 (93.73\% vs.\ 91.60\%), but the difference is not significant (\(p=0.089\)).

We then examine the effect of BAC fine-tuning on the Qwen3 backbone. The best configuration identified by our fine-tuning sweep uses a learning rate of \(5\times10^{-5}\), LoRA rank \(r=16\), scaling factor \(\alpha=32\), and dropout \(0.05\). This configuration shows stable training and validation convergence across all three model scales, with full optimization details and learning curves provided in Appendix~\ref{app:finetuning}. 

BAC substantially improves the corresponding Qwen3 baseline at every scale. Although BAC is trained only on the captioning task, it generalizes strongly to person-centric QA without QA-specific supervision. QA accuracy increases from 68.90\% to 80.33\% at 2B, from 79.73\% to 90.13\% at 4B, and from 87.70\% to 93.20\% at 8B; all three improvements are statistically significant (\(p<0.001\)). The gains are especially large for UNIDENTIFIED questions at 2B and 4B. Figure~\ref{fig:bac_qualitative_examples}(a) qualitatively illustrates this improvement, with BAC correctly predicting \texttt{UNIDENTIFIED} across all three model scales while the corresponding Qwen3-VL baselines fail. Named accuracy also remains high, indicating that adaptation improves the model's ability to distinguish grounded identities from people whose identity is not provided.

Captioning follows the same trend: BAC improves overall quality and identity accuracy at all three scales while maintaining a low hallucination rate. BAC-2B already slightly exceeds the strongest 4B base model in QA (80.33\% vs.\ 79.73\%), while BAC-4B surpasses all evaluated 8--9B base models (90.13\% vs.\ 87.70\% for the strongest baseline). The relative improvement decreases as model size grows, suggesting that task-specific adaptation is particularly beneficial for smaller models, where identity grounding is more challenging.

\subsection{Comparison with Frontier VLMs}

Table~\ref{tab:frontier_qa} compares BAC with substantially larger open-weight and proprietary frontier VLMs. GPT-5.6 Sol achieves the highest overall performance, reaching 97.07\% with thinking and 96.30\% without thinking; the difference between the two settings is small but statistically significant (\(p=0.007\)). BAC-8B reaches 93.20\% overall, with balanced performance on Named (93.88\%) and UNIDENTIFIED (91.68\%) questions. Although it remains significantly below GPT-5.6 Sol (\(p<0.001\)), BAC-8B significantly outperforms Qwen3-VL-235B-A22B (86.10\%), Gemini 2.5 Flash (82.63\%), and Claude Sonnet 4.6 (76.10\%), all with \(p<0.001\). 

This advantage is already visible at smaller scales. BAC-4B reaches 90.13\% overall accuracy and significantly surpasses the much larger Qwen3-VL-235B-A22B by 4.03 percentage points (\(p<0.001\)), while achieving 96.38\% accuracy on Named questions. BAC-2B also reaches 80.33\%, significantly outperforming Claude Sonnet 4.6 (\(p<0.001\)), although it remains below Gemini 2.5 Flash and Qwen3-VL-235B-A22B. Overall, these results show that BAC substantially narrows the gap to frontier proprietary models while remaining compact, open-weight, and self-hostable. Figure~\ref{fig:bac_qualitative_examples} illustrates representative BAC successes and failures: (a) correct UNIDENTIFIED prediction, (b) correct named-person identification, and (c) a challenging failure where only GPT-5.6 Sol succeeds. Additional qualitative examples and frontier-model failures are provided in Appendix~\ref{app:qualitative}.

Category-wise results show that BAC-8B is particularly strong on appearance and object questions, reaching 95.58\% and 93.99\%, respectively. Its lowest performance is observed on position questions (87.54\%), followed by interaction questions (91.28\%), suggesting that spatial and relational reasoning remain more challenging than attribute-based recognition. A detailed breakdown across all BAC model scales and frontier Video-MLLMs is provided in Appendix~\ref{app:category_qa}. 

\begin{table}[h]
\centering
\footnotesize
\setlength{\tabcolsep}{4pt}
\renewcommand{\arraystretch}{0.95}

\caption{
Benchmark comparison across model families and sizes with the FVB+FTC grounding method.
Captioning is evaluated using overall quality, identity accuracy, and
hallucination rate. QA performance is reported over all questions,
separately for Named and  UNIDENTIFIED (UNID).
}
\label{tab:model_benchmark}

\resizebox{0.85\textwidth}{!}{%
\begin{tabular}{llccc|ccc}
\toprule
& &
\multicolumn{3}{c|}{\textbf{Captioning}} &
\multicolumn{3}{c}{\textbf{Person-Centric QA}} \\
\cmidrule(lr){3-5}
\cmidrule(lr){6-8}

\textbf{Size} &
\textbf{Model} &
\textbf{Overall} $\uparrow$ &
\textbf{Identity} $\uparrow$ &
\textbf{Halluc.} $\downarrow$ &
\textbf{Overall} $\uparrow$ &
\textbf{Named} $\uparrow$ &
\textbf{UNID} $\uparrow$ \\
\midrule

\multirow{3}{*}{\textbf{2B}}
& \textbf{BAC (ours)}
& \textbf{8.09} & \textbf{85.33} & 3.20
& \textbf{80.33} & \textbf{95.90} & \textbf{45.46} \\

& Qwen3
& 7.61 & 76.53 & \textbf{3.07}
& 68.90 & 92.48 & 16.09 \\

& InternVL3.5
& 5.59 & 24.86 & 4.48
& 63.90 & 91.18 & 2.81 \\
\midrule

\multirow{5}{*}{\textbf{4B}}
& \textbf{BAC (ours)}
& \textbf{8.48} & \textbf{92.40} & \textbf{0.80}
& \textbf{90.13} & \textbf{96.38} & \textbf{76.13} \\

& Qwen3
& 8.32 & 88.80 & 1.07
& 79.73 & 93.11 & 49.78 \\

& InternVL3.5
& 7.49 & 76.93 & 6.80
& 74.87 & 70.38 & 84.99 \\

& MiniCPM-V
& 6.61 & 60.93 & 15.47
& 69.60 & 80.23 & 45.79 \\

& Ovis2
& 6.40 & 47.87 & 10.80
& 65.27 & 71.12 & 52.16 \\
\midrule

\multirow{6}{*}{\textbf{8--9B}}
& \textbf{BAC (ours)}
& \textbf{8.59} & \textbf{94.03} & 1.06
& \textbf{93.20} & \textbf{93.88} & \textbf{91.68} \\

& Qwen3
& 8.45 & 91.60 & \textbf{0.93}
& 87.70 & 88.04 & 86.93 \\

& Ovis2.5 9B
& 8.39 & 93.73 & 1.87
& 83.10 & 85.39 & 77.97 \\

& InternVL3.5
& 7.68 & 81.73 & 6.27
& 79.00 & 85.28 & 65.12 \\

& Eagle2.5
& 7.42 & 74.13 & 4.80
& 77.63 & 79.17 & 74.19 \\

& Molmo2
& 5.66 & 52.40 & 28.13
& 69.17 & 63.36 & 82.18 \\

\bottomrule
\end{tabular}%
}
\end{table}

\begin{table*}[h]
\centering
\footnotesize
\setlength{\tabcolsep}{4pt}
\renewcommand{\arraystretch}{0.95}

\caption{
Person-centric QA performance and model accessibility.
Mean denotes the average of Named and UNIDENTIFIED (UNID) accuracy.
API costs are reported in USD per million input/output tokens;
All models use the FVB+FTC grounding method.
}
\label{tab:frontier_qa}

\resizebox{\textwidth}{!}{%
\begin{tabular}{lccccccc}
\toprule
\textbf{Model} &
\textbf{Overall} &
\textbf{Named} &
\textbf{UNID} &
\textbf{Mean} &
\textbf{Clip all-correct} &
\textbf{Open-weight} &
\textbf{Cost (\$/1M in/out)} \\
\midrule

GPT-5.6 Sol (think)
& \textbf{97.07}
& 97.06
& 97.08
& \textbf{97.07}
& \textbf{90.13}
& No
& 4.00 / 20.00 \\

GPT-5.6 Sol (no think)
& 96.30
& 96.14
& 96.65
& 96.40
& 87.60
& No
& 4.00 / 20.00 \\

\textbf{BAC-8B (ours)}
& \textbf{93.20}
& 93.88
& 91.68
& 92.78
& 76.93
& Yes
& Self-hosted \\

\textbf{BAC-4B (ours)}
& \textbf{90.13}
& \textbf{96.38}
& 76.13
& 86.26
& 66.40
& Yes
& Self-hosted \\

Qwen3-VL-235B-A22B
& 86.10
& 80.67
& \textbf{98.27}
& 89.47
& 60.93
& Yes
& 0.287 / 1.147 \\

Gemini 2.5 Flash
& 82.63
& 75.80
& 97.95
& 86.87
& 51.60
& No
& 0.30 / 2.50 \\

\textbf{BAC-2B (ours)}
& \textbf{80.33}
& 95.90
& 45.46
& 70.68
& 43.73
& Yes
& Self-hosted \\

Claude Sonnet 4.6
& 76.10
& 66.44
& 97.73
& 82.09
& 41.87
& No
& 3.00 / 15.00 \\

\bottomrule
\end{tabular}%
}
\end{table*}

\section{Conclusion}
We introduced a framework for identity-aware video understanding that combines automatic character identification with explicit spatial grounding for video captioning and person-centric question answering. Using LSMDC v2, we constructed a manually verified benchmark of 750 captioned clips and 3,000 questions, and systematically evaluated five identity-grounding representations across multiple model scales and families. Our results show that explicit visual grounding substantially improves identity association compared with textual coordinates alone, with face-based visual grounding combined with textual coordinates providing the most consistent performance across scales. Building on this formulation, we introduced BAC through LoRA-based fine-tuning of Qwen models at 2B, 4B, and 8B scales. BAC-8B achieves 93.20\% QA accuracy and remains competitive with substantially larger frontier Video-MLLMs, while maintaining strong performance across question categories. Overall, the results show that explicitly communicating who is where, together with task-specific adaptation, provides an effective approach to improving identity-aware video understanding without requiring changes to the underlying Video-MLLM architecture.

\section{Ethical Considerations}
Our approach combines face recognition, temporal tracking, and spatial grounding to associate known identities with their appearances, actions, and interactions across video frames. Although we study this capability only in the controlled setting of movie understanding, using predefined cast identities and movie clips, similar techniques could potentially be adapted to identify and track individuals in real-world footage, raising concerns related to privacy, surveillance, and unauthorized profiling. Our work is intended to advance character-centric video understanding rather than person identification in unconstrained real-world environments, and we do not evaluate or advocate its use for surveillance or monitoring applications. We therefore encourage future applications of identity-aware video understanding to consider appropriate consent, privacy protections, and restrictions on the use of biometric identity information.

\section{Acknowledgments}
\label{sec:acknowledgments}

This project was provided with computing AI and storage resources by GENCI at IDRIS thanks to the grant 20XX-AD011017029 on the supercomputer Jean Zay’s H100 partition.

\bibliography{iclr2027_conference}
\bibliographystyle{iclr2027_conference}

\appendix
\section*{Appendix}

\section{Estimating a Person Bounding Box from a Face Detection}
\label{app:pbbox}

To obtain person-level spatial grounding without requiring an additional person detector, we estimate a full-person bounding box directly from each detected face. Given a face bounding box \(B_{\text{face}}=[x_f,y_f,w_f,h_f]\), we exploit approximate human body proportions to construct an estimated person box \(B_{\text{body}}=[x_b,y_b,w_b,h_b]\). Specifically, the body height and width are obtained by scaling the detected face dimensions,

\[
h_b = K_h h_f,
\qquad
w_b = K_w w_f,
\]

where \(K_h\) and \(K_w\) control the expected body-to-face height and width ratios, respectively. We assume that the body is approximately horizontally aligned with the face. We therefore compute the horizontal center of the face,

\[
x_{c,f}=x_f+\frac{w_f}{2},
\]

and center the estimated body box around this position,

\[
x_b=x_{c,f}-\frac{w_b}{2}.
\]

Vertically, the person box begins slightly above the detected face in order to include the top of the head,

\[
y_b=y_f-K_{\mathrm{offset}}h_f,
\]

where \(K_{\mathrm{offset}}\) controls the upward extension. The resulting box therefore preserves the spatial location of the detected face while extending it according to approximate human proportions to cover the person's visible body. This procedure is heuristic rather than a learned body-detection method and provides the estimated person regions used in our person-box grounding representations.

\paragraph{Implementation parameters.}
We use the same fixed parameters for all experiments. Specifically, we empirically set the body-height scaling factor to \(K_h=6.5\), the body-width scaling factor to \(K_w=2.5\), and the vertical offset to \(K_{\mathrm{offset}}=0.15\). Thus, the estimated person box has a height of \(6.5h_f\) and a width of \(2.5w_f\), and starts \(0.15h_f\) above the detected face box. These parameters are applied uniformly to all detected faces without additional person detection or per-frame adaptation.

\section{Effect of Face and Person Size}
\label{app:effect_size}

As shown in the main results, the relative effectiveness of FVB+FTC and PVB+PTC depends partly on target size. When the face is very small, the full-person box can provide a more informative spatial cue. In Figure~\ref{fig:example66_face_vs_person}, the target face occupies only 2,222 px$^2$ with a height of 52 px; PVB+PTC correctly identifies the target across all Qwen3-VL sizes, whereas FVB+FTC fails.

\begin{figure}[h]
    \centering

    \includegraphics[width=\textwidth]
    {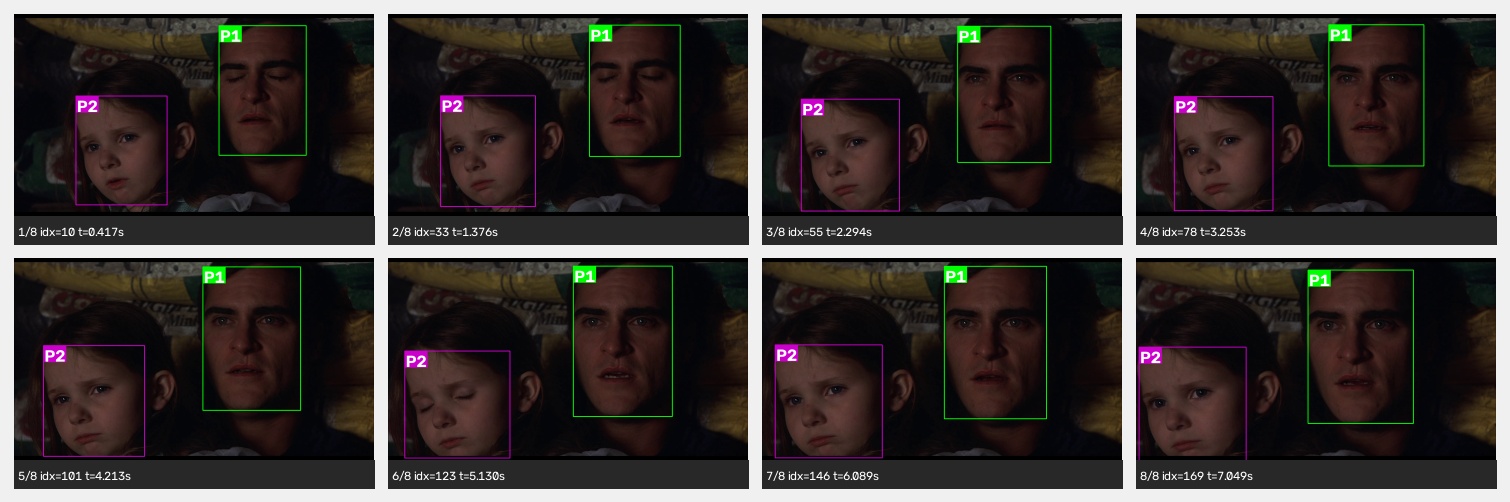}

    \vspace{-1mm}
    {\small\textbf{(a) Face grounding}}

    \vspace{2mm}

    \includegraphics[width=\textwidth]
    {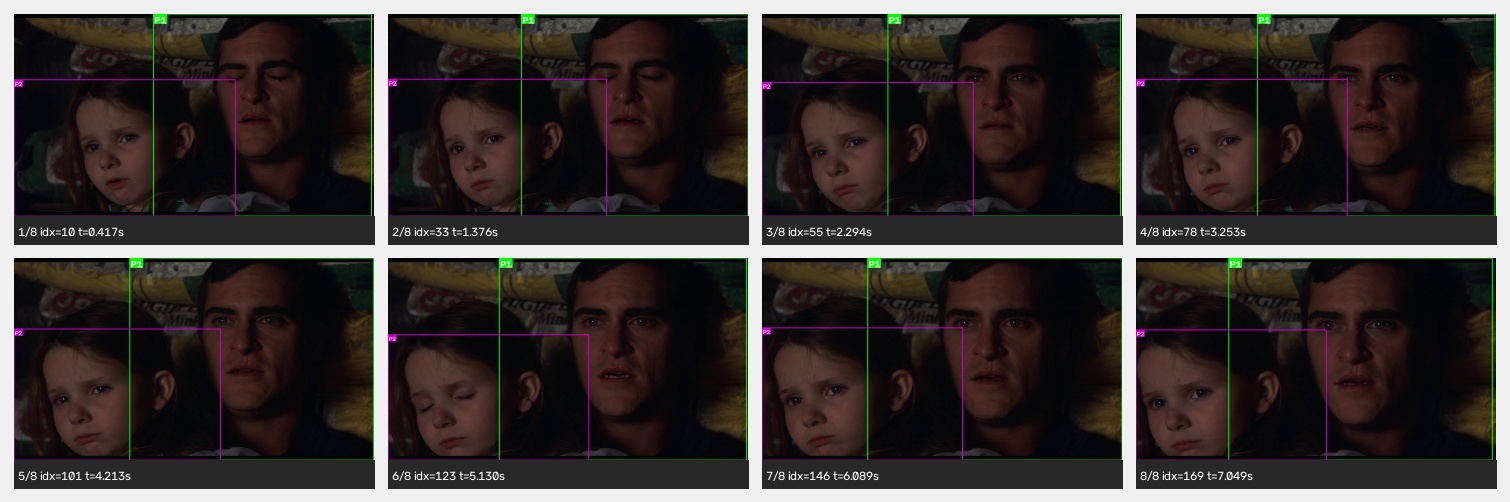}

    \vspace{-1mm}
    {\small\textbf{(b) Person grounding}}

    \vspace{2mm}

    \begin{minipage}{0.98\textwidth}
    \small

    \textbf{Identity mapping:}
    \textcolor{green!60!black}{P1} $\leftrightarrow$
    \textcolor{green!60!black}{Joaquin};
    \textcolor{magenta!80!black}{P2} $\leftrightarrow$
    \textcolor{magenta!80!black}{Abigail}.

    \vspace{1mm}

    \textbf{Q2 [Action]:}
    \textit{Who opens their eyes with a frightened expression?}
    \hspace{0.5em}
    \textbf{GT:} \textcolor{green!60!black}{Joaquin}.
    \hspace{0.5em}
    \textbf{P1 face size:}
    391,897 px$^2$ (760.5 px height).

    \vspace{1mm}

    \centering
\begin{tabular}{c|cc}
    \toprule
    \textbf{Model} & \textbf{FVB+FTC} & \textbf{PVB+PTC} \\
    \midrule
    Qwen3-VL 2B &
    $\checkmark$ \textcolor{green!60!black}{Joaquin} &
    $\times$ \textcolor{magenta!80!black}{Abigail} \\

    Qwen3-VL 4B &
    $\checkmark$ \textcolor{green!60!black}{Joaquin} &
    $\checkmark$ \textcolor{green!60!black}{Joaquin} \\

    Qwen3-VL 8B &
    $\checkmark$ \textcolor{green!60!black}{Joaquin} &
    $\checkmark$ \textcolor{green!60!black}{Joaquin} \\
    \bottomrule
\end{tabular}

    \end{minipage}

    \caption{
    Qualitative comparison of face- and person-level grounding for an
    action question. The target character
    \textcolor{green!60!black}{P1} (\textcolor{green!60!black}{Joaquin})
    has a very large face area of 391,897 px$^2$ and a face height of 760.5 px.
    FVB+FTC grounding method correctly identifies
    \textcolor{green!60!black}{Joaquin} across all three model sizes.
    PVB+PTC grounding method predicts
    \textcolor{magenta!80!black}{Abigail} at 2B, but correctly identifies
    \textcolor{green!60!black}{Joaquin} at 4B and 8B.
    }
    \label{fig:example67_face_vs_person}
\end{figure}

Conversely, when the face is large and visually informative, FVB+FTC can be more reliable. In Figure~\ref{fig:example67_face_vs_person}, the face occupies 391,897 px$^2$ with a height of 760.5 px; FVB+FTC succeeds across all model sizes, while PVB+PTC fails for Qwen3-VL 2B. The full-person boxes also overlap in this example, introducing visual clutter that can weaken the identity--location association.

These examples support the quantitative results: PVB+PTC is particularly useful for very small faces, whereas FVB+FTC becomes more reliable as face size increases. For already informative faces, full-person boxes may add little benefit and can even interfere with grounding when they overlap.

\definecolor{p4yellow}{RGB}{255, 140, 0}

\begin{figure}[h]
    \centering

    \includegraphics[width=\textwidth]
    {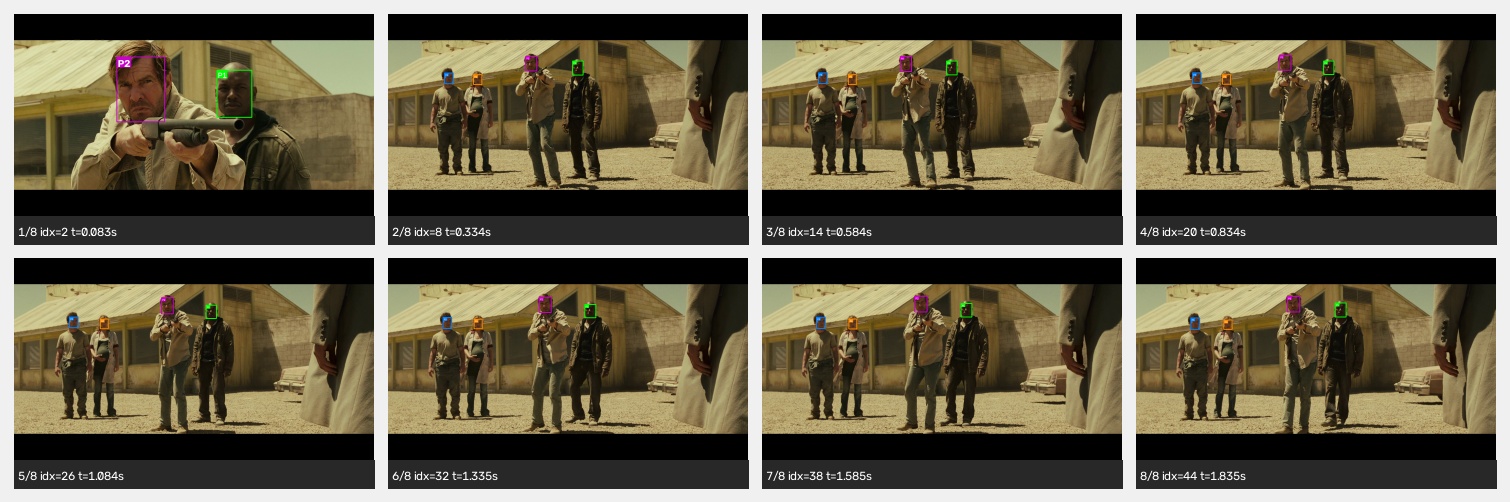}

    \vspace{-1mm}
    {\small\textbf{(a) Face grounding}}

    \vspace{2mm}

    \includegraphics[width=\textwidth]
    {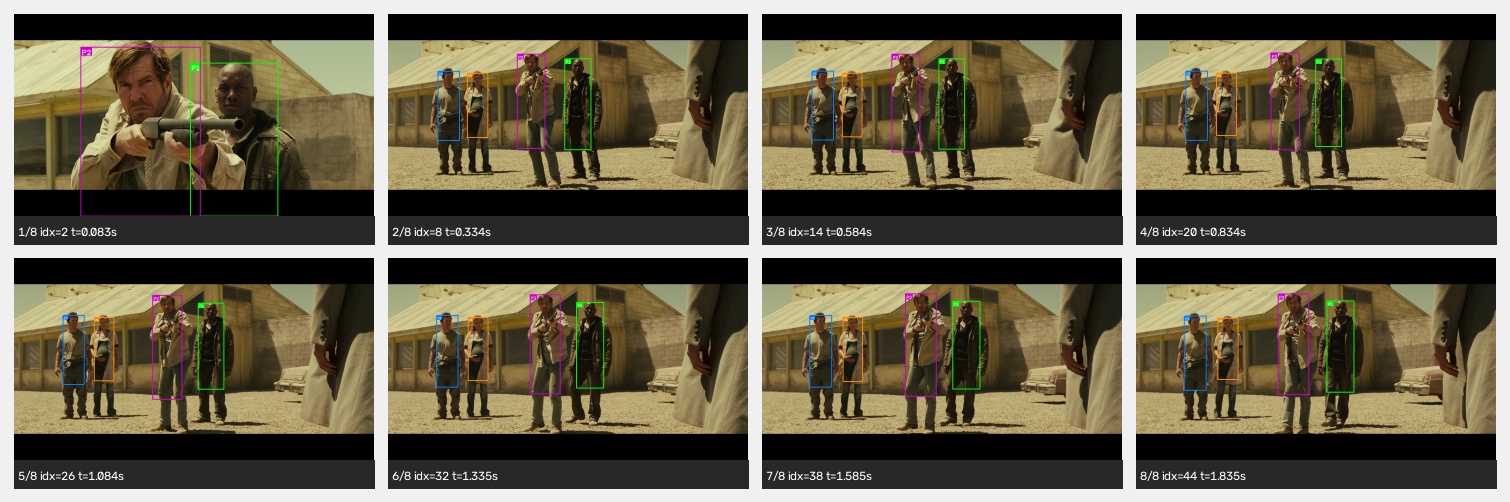}

    \vspace{-1mm}
    {\small\textbf{(b) Person grounding}}

    \vspace{2mm}

    \begin{minipage}{0.98\textwidth}
    \small

    \textbf{Identity mapping:}
    \textcolor{green!60!black}{P1} $\leftrightarrow$
    \textcolor{green!60!black}{Tyrese Gibson};
    \textcolor{magenta!80!black}{P2} $\leftrightarrow$
    \textcolor{magenta!80!black}{Dennis Quaid};
    \textcolor{blue}{P3} $\leftrightarrow$
    \textcolor{blue}{Lucas Black};
    \textcolor{p4yellow}{P4} $\leftrightarrow$
    \textcolor{p4yellow}{Adrianne Palicki}.

    \vspace{1mm}

    \textbf{Q2 [Appearance]:}
    \textit{Who is pregnant?}
    \hspace{0.5em}
    \textbf{GT:} \textcolor{p4yellow}{Adrianne Palicki}.
    \hspace{0.5em}
    \textbf{P4 face size:}
    2,222 px$^2$ (52 px height).

    \vspace{1mm}

    \centering
\begin{tabular}{c|cc}
    \toprule
    \textbf{Model} & \textbf{FVB+FTC} & \textbf{PVB+PTC} \\
    \midrule
    Qwen3-VL 2B &
    $\times$ \textcolor{blue}{Lucas} &
    $\checkmark$ \textcolor{p4yellow}{Adrianne} \\

    Qwen3-VL 4B &
    $\times$ \textcolor{blue}{Lucas} &
    $\checkmark$ \textcolor{p4yellow}{Adrianne} \\

    Qwen3-VL 8B &
    $\times$ \texttt{UNIDENTIFIED} &
    $\checkmark$ \textcolor{p4yellow}{Adrianne} \\
    \bottomrule
\end{tabular}

    \end{minipage}

    \caption{
    Qualitative comparison of face- and person-level grounding for a
    small-face example. The target character
    \textcolor{p4yellow}{P4} (\textcolor{p4yellow}{Adrianne Palicki})
    has a very small face area of 2,222 px$^2$ and a face height of 52 px.
    PVB+PTC grounding method correctly identifies
    \textcolor{p4yellow}{Adrianne Palicki} across all model sizes,
    whereas FVB+FTC grounding method predicts
    \textcolor{blue}{Lucas Black} at 2B and 4B and
    \texttt{UNIDENTIFIED} at 8B.
    }
    \label{fig:example66_face_vs_person}
\end{figure}

\clearpage
\section{BAC Fine-Tuning and Model Scaling}
\label{app:finetuning}
\paragraph{LoRA fine-tuning.}
Our objective is to fine-tune Qwen3-VL base models at different scales on the training data we build as described in Section~\ref{subsec:Caption_and_Question Generation}. We consider the 2B, 4B, and 8B variants and perform parameter-efficient adaptation using LoRA. We refer to the resulting fine-tuned models as \textbf{BAC-2B}, \textbf{BAC-4B}, and \textbf{BAC-8B}. The adapters are applied only to the language-model projection layers, while the vision encoder and multimodal projector remain frozen.

\begin{figure}[h]
    \centering
    
    \begin{subfigure}{0.85\textwidth}
        \centering
        \includegraphics[width=\linewidth]{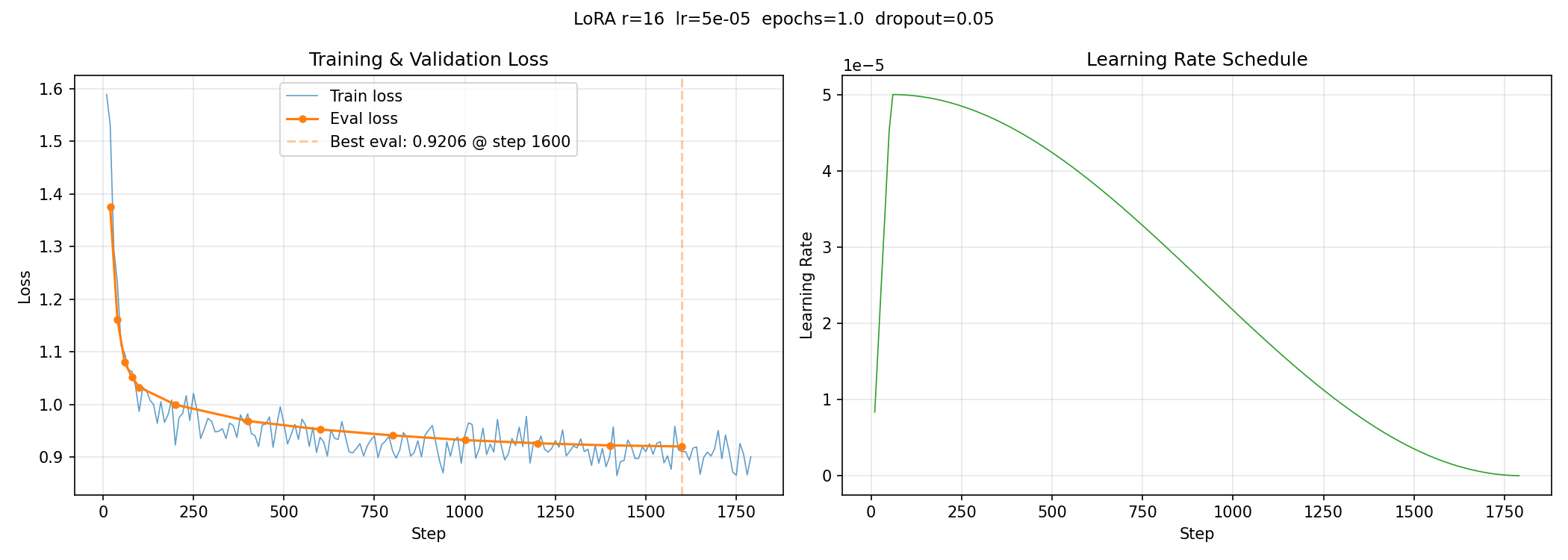}
        \caption{BAC-2B}
        \label{fig:bac_2b_training_curve}
    \end{subfigure}
    
    \vspace{0.5em}
    
    \begin{subfigure}{0.85\textwidth}
        \centering
        \includegraphics[width=\linewidth]{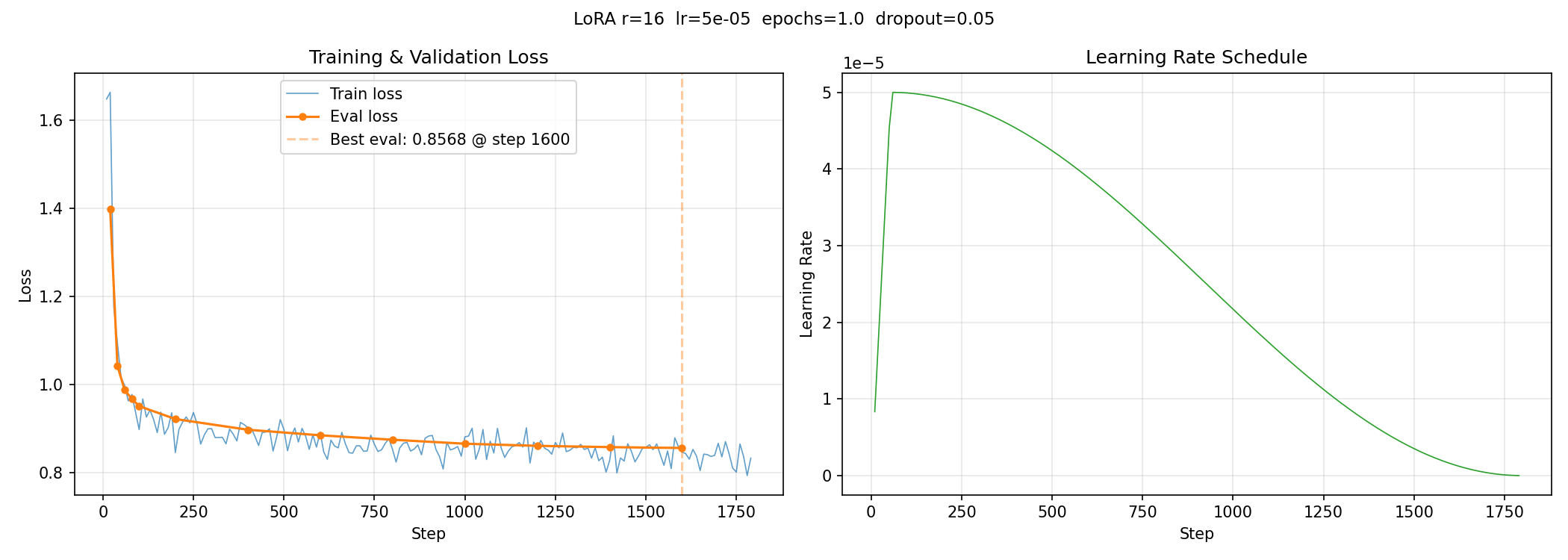}
        \caption{BAC-4B}
        \label{fig:bac_4b_training_curve}
    \end{subfigure}
    
    \vspace{0.5em}
    
    \begin{subfigure}{0.85\textwidth}
        \centering
        \includegraphics[width=\linewidth]{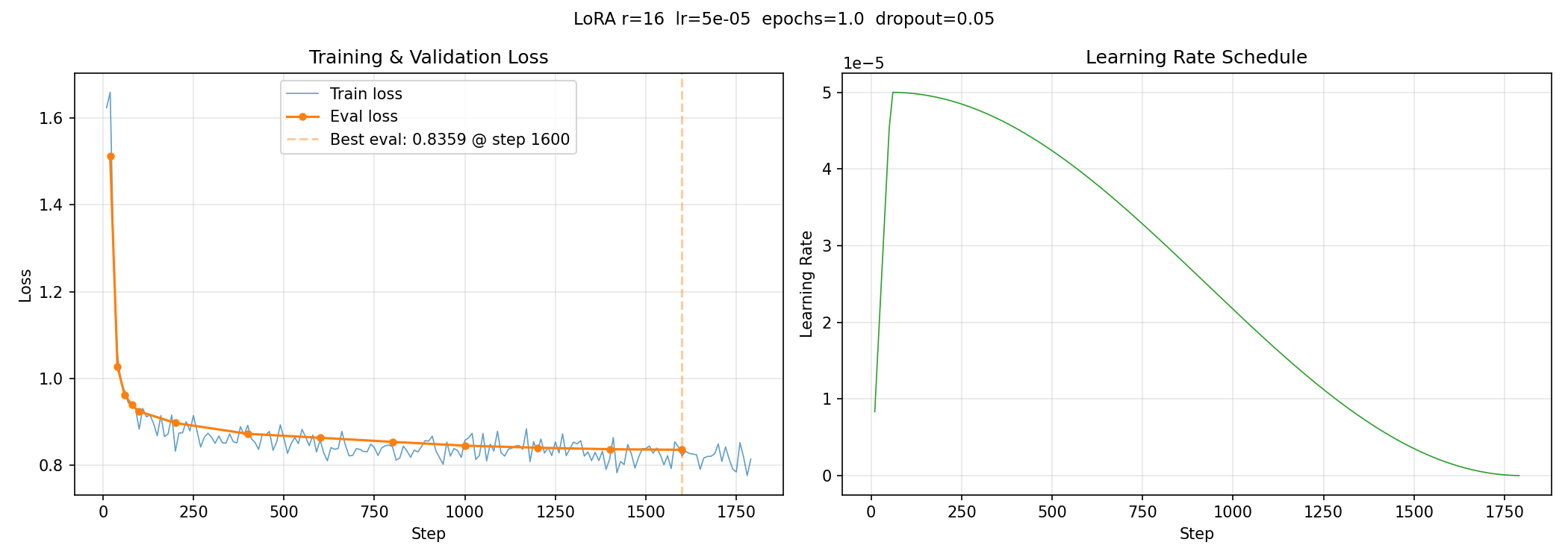}
        \caption{BAC-8B}
        \label{fig:bac_8b_training_curve}
    \end{subfigure}
    
    \caption{Training dynamics of the BAC models at three scales. For each model, we report the training loss, evaluation loss, and learning-rate schedule during LoRA fine-tuning. The dashed vertical line indicates the checkpoint with the best evaluation loss.}
    \label{fig:bac_training_dynamics}
\end{figure}

To determine an effective LoRA configuration, we conduct a five-setting hyperparameter sweep for one epoch, varying the learning rate, LoRA rank \(r\), scaling factor \(\alpha\), and dropout. The best configuration is obtained  with a learning rate of \(5\times10^{-5}\), \(r=16\), \(\alpha=32\), and a dropout of \(0.05\). At this rank, LoRA introduces \(17.4\)M trainable parameters for BAC-2B, \(33.0\)M for BAC-4B, and \(43.6\)M for BAC-8B. Figure~\ref{fig:bac_training_dynamics} shows the training and validation losses for the 2B, 4B, and 8B models using the best configuration, with both curves reported together for each model size. For each BAC variant, the checkpoint achieving the lowest validation loss is used for inference.

\clearpage
\section{Category-Wise Person-Centric QA Performance}
\label{app:category_qa}
\begin{figure}[h]
    \centering

    \includegraphics[width=\textwidth]{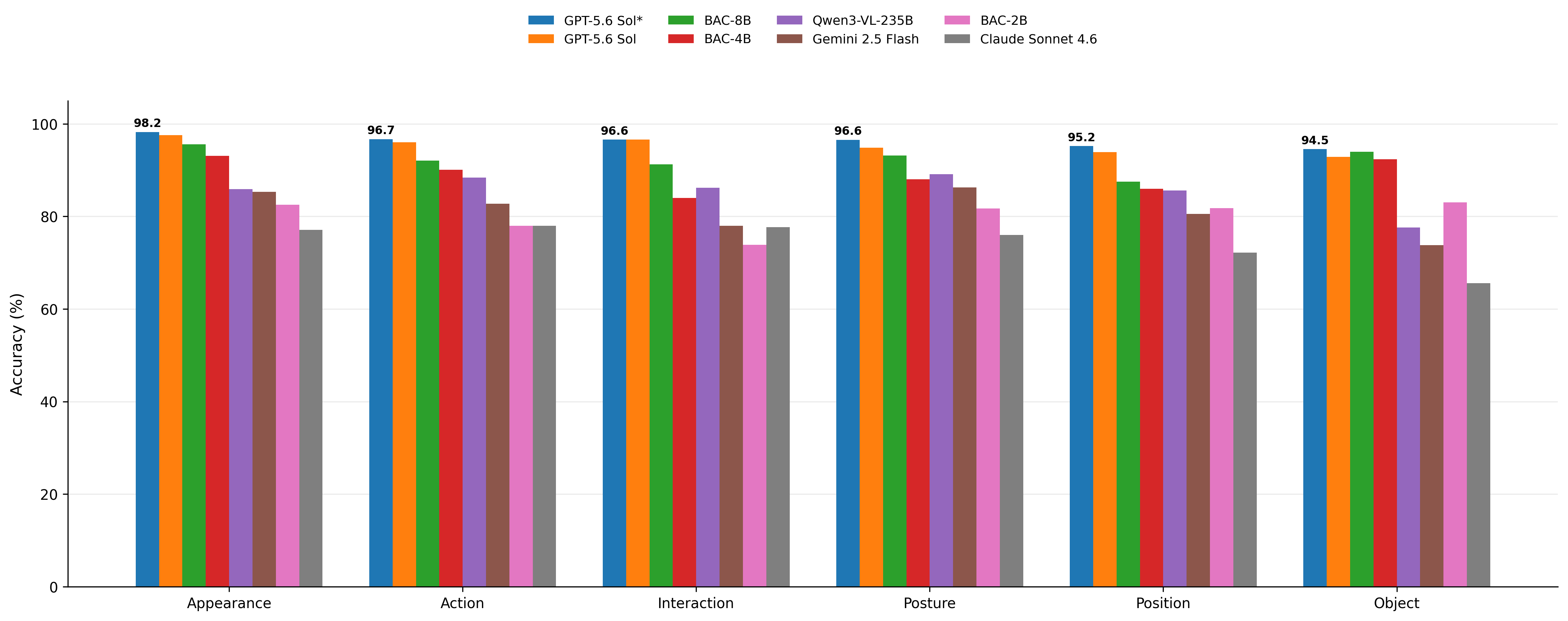}

    \vspace{0.3cm}

    \includegraphics[width=\textwidth]{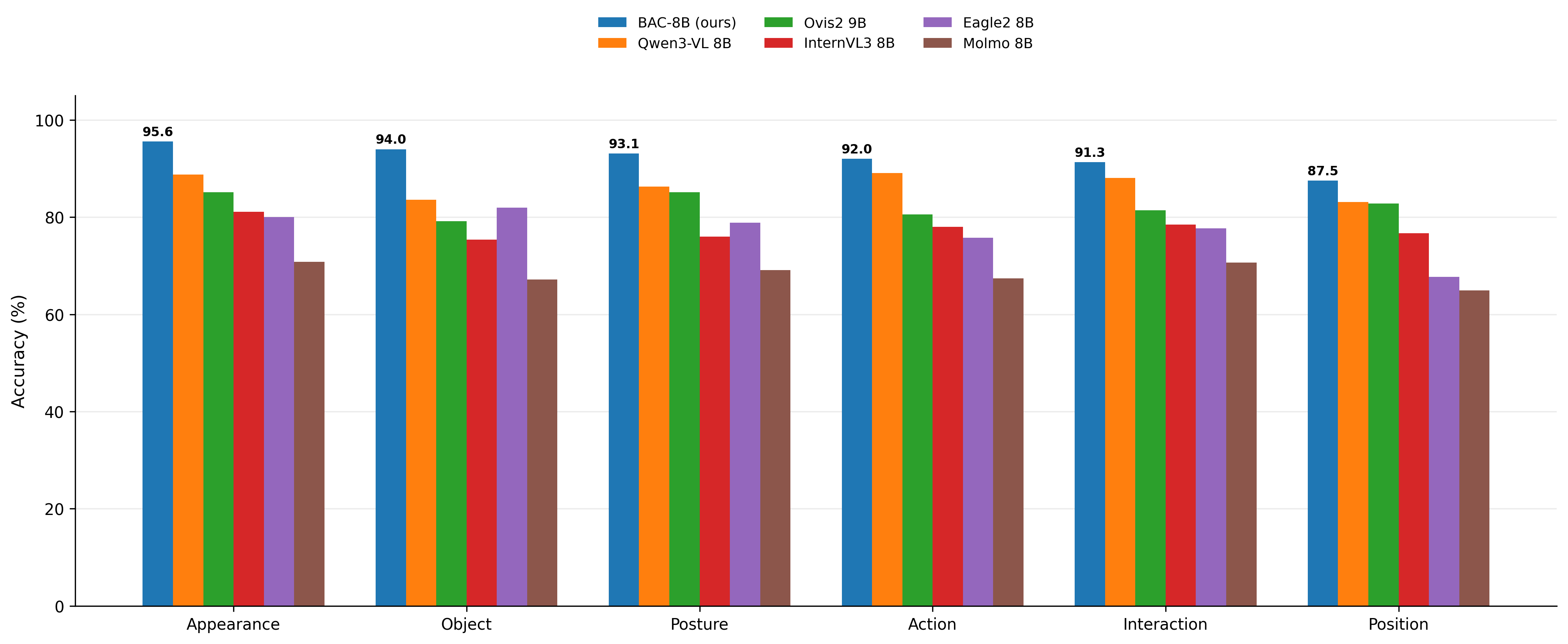}

    \caption{
    Person-centric QA accuracy across question categories.
    Top: comparison with frontier models, with categories ordered by
    GPT-5.6 Sol* performance. Bottom: comparison between BAC-8B and
    other 8--9B models, with categories ordered by BAC-8B performance.
    GPT-5.6 Sol* denotes GPT-5.6 Sol with thinking enabled.
    }
    \label{fig:qa_category_comparison}
\end{figure}

Figure~\ref{fig:qa_category_comparison} provides a category-wise breakdown of person-centric QA performance across both frontier and similarly sized models. Overall, BAC-8B maintains strong performance across all question types, indicating that task-specific adaptation benefits a broad range of identity-aware reasoning categories.

Among frontier models, GPT-5.6 Sol thinking achieves the highest accuracy across all categories, while BAC-8B remains competitive, with its strongest absolute results on appearance, object, and posture questions and its lowest performance on position. BAC-4B follows a similar pattern, whereas BAC-2B substantially larger variation across categories. Among 8--9B models, BAC-8B achieves the highest accuracy in every category and consistently outperforms its Qwen3-VL 8B base, with the largest gains over the base model observed for object, appearance, and posture questions.
\label{app:categories_results}

\clearpage

\begingroup
\small
\setlength{\textwidth}{0.9\textwidth}
\setlength{\tabcolsep}{2pt}
\renewcommand{\arraystretch}{0.9}

\renewcommand{\caption}[1]{}

\input{add_results_latex.tex}

\endgroup

\end{document}

%% file: math_commands.tex
\usepackage{amsmath,amsfonts,bm}

\def\eqref#1{equation~\ref{#1}}

\def\1{\bm{1}}

\DeclareMathAlphabet{\mathsfit}{\encodingdefault}{\sfdefault}{m}{sl}
\SetMathAlphabet{\mathsfit}{bold}{\encodingdefault}{\sfdefault}{bx}{n}



%% file: add_results_latex.tex
\definecolor{p4orange}{RGB}{230,126,34}

\section{Additional Qualitative Results}
\label{app:qualitative}
{%
\centering

    \includegraphics[width=0.98\textwidth]
    {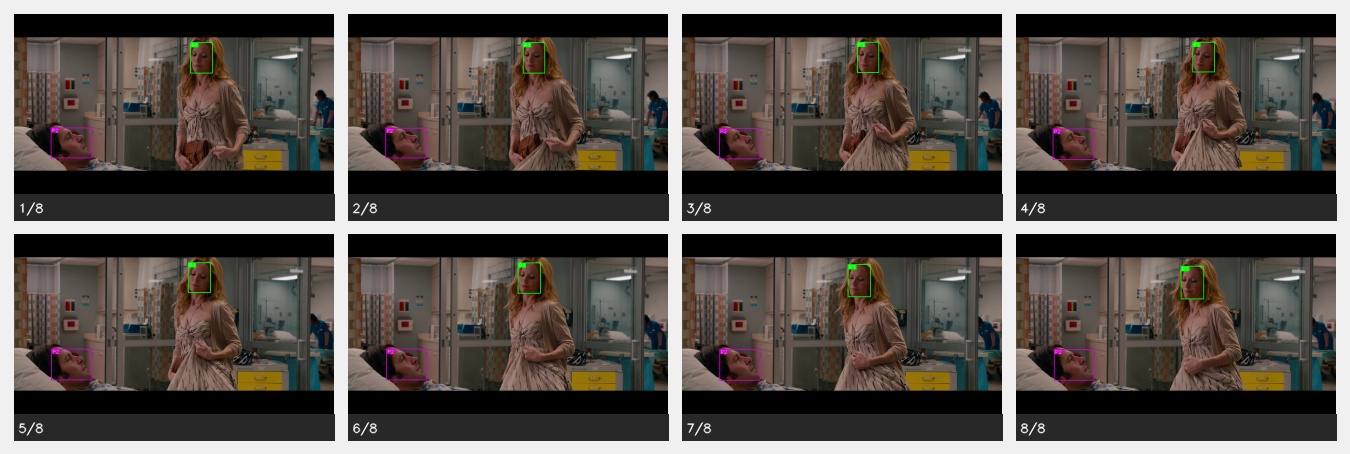}

    \vspace{0.3mm}

    \begin{minipage}{0.98\textwidth}
    \footnotesize

    \textbf{(a) Identity mapping:}
    \textcolor{green!60!black}{P1} $\leftrightarrow$
    \textcolor{green!60!black}{Leslie Mann};
    \textcolor{magenta!80!black}{P2} $\leftrightarrow$
    \textcolor{magenta!80!black}{Paul Rudd}.

    \vspace{0.2mm}

    \textbf{Q2 [Posture]:}
    \textit{Who is lying in a hospital bed?}
    \hspace{0.6em}
    \textbf{GT:}
    \textcolor{magenta!80!black}{Paul Rudd}
    (\textcolor{magenta!80!black}{P2}).

    \textbf{Q3 [Action]:}
    \textit{Who gathers her dress as she walks past the hospital bed?}
    \hspace{0.6em}
    \textbf{GT:}
    \textcolor{green!60!black}{Leslie Mann}
    (\textcolor{green!60!black}{P1}).

    \vspace{0.3mm}

    \centering
    \scriptsize
    \setlength{\tabcolsep}{2.5pt}
    \renewcommand{\arraystretch}{0.90}

    \begin{tabular}{
        @{}lccc@{\hspace{6pt}}
        lccc@{\hspace{6pt}}
        lccc@{}
    }
        \toprule
        \textbf{Model} & \textbf{Q2} & \textbf{Q3} & &
        \textbf{Model} & \textbf{Q2} & \textbf{Q3} & &
        \textbf{Model} & \textbf{Q2} & \textbf{Q3} & \\
        \midrule

        \textbf{BAC-2B (ours)} &
        \textcolor{green!60!black}{$\checkmark$} &
        \textcolor{green!60!black}{$\checkmark$} & &

        \textbf{BAC-4B (ours)} &
        \textcolor{green!60!black}{$\checkmark$} &
        \textcolor{green!60!black}{$\checkmark$} & &

        \textbf{BAC-8B (ours)} &
        \textcolor{green!60!black}{$\checkmark$} &
        \textcolor{green!60!black}{$\checkmark$} & \\

        Qwen3-VL 2B &
        \textcolor{green!60!black}{$\checkmark$} &
        \textcolor{green!60!black}{$\checkmark$} & &

        Qwen3-VL 4B &
        \textcolor{green!60!black}{$\checkmark$} &
        \textcolor{green!60!black}{$\checkmark$} & &

        Qwen3-VL 8B &
        \textcolor{green!60!black}{$\checkmark$} &
        \textcolor{green!60!black}{$\checkmark$} & \\

        \midrule

        GPT-5.6 Sol &
        Leslie \textcolor{red}{$\times$} &
        Paul \textcolor{red}{$\times$} & &

        GPT-5.6 Sol$^{*}$ &
        Leslie \textcolor{red}{$\times$} &
        Paul \textcolor{red}{$\times$} & &

        Claude Sonnet 4.6 &
        Leslie \textcolor{red}{$\times$} &
        Paul \textcolor{red}{$\times$} & \\

        Qwen3-VL 235B &
        \textcolor{green!60!black}{$\checkmark$} &
        \textcolor{green!60!black}{$\checkmark$} & &

        Gemini 2.5 Flash &
        \textcolor{green!60!black}{$\checkmark$} &
        \textcolor{green!60!black}{$\checkmark$} & &

        & & & \\

        \bottomrule
    \end{tabular}

    \end{minipage}

    \vspace{1.2mm}

    \includegraphics[width=0.98\textwidth]
    {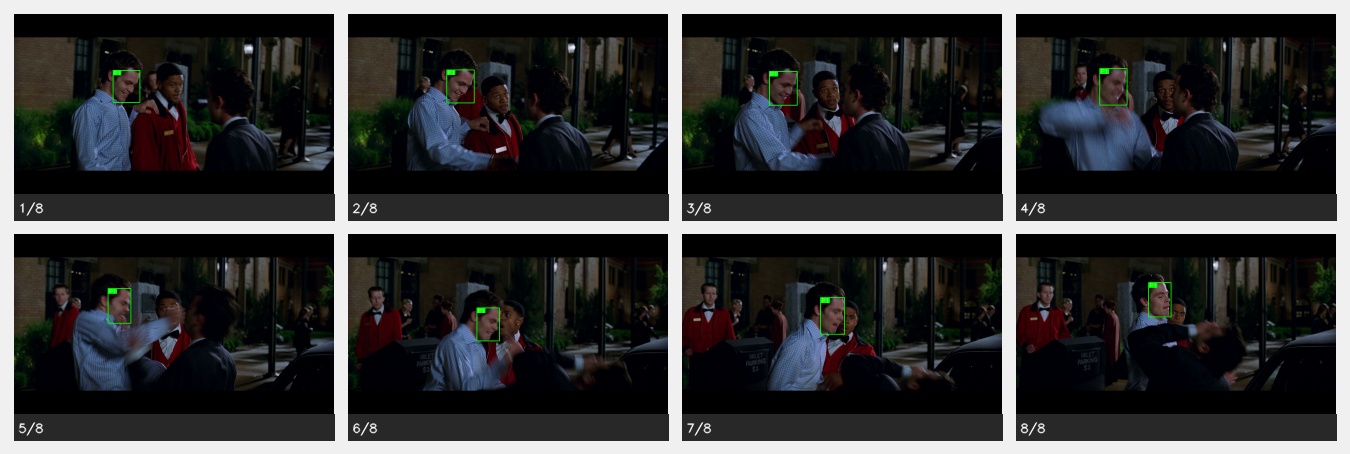}

    \vspace{0.3mm}

    \begin{minipage}{0.98\textwidth}
    \footnotesize

    \textbf{(b) Identity mapping:}
    \textcolor{green!60!black}{P1} $\leftrightarrow$
    \textcolor{green!60!black}{Chris Pine}.

    \vspace{0.2mm}

    \textbf{Q3 [Interaction]:}
    \textit{Who is being punched in the face?}
    \hspace{0.6em}
    \textbf{GT:} \texttt{UNIDENTIFIED}.

    \vspace{0.3mm}

    \centering
    \scriptsize
    \setlength{\tabcolsep}{2.5pt}
    \renewcommand{\arraystretch}{0.90}

    \begin{tabular}{
        @{}lcc@{\hspace{8pt}}
        lcc@{\hspace{8pt}}
        lcc@{}
    }
        \toprule
        \textbf{Model} & \textbf{Pred.} & &
        \textbf{Model} & \textbf{Pred.} & &
        \textbf{Model} & \textbf{Pred.} & \\
        \midrule

        \textbf{BAC-2B (ours)} &
        Chris &
        \textcolor{red}{$\times$} &

        \textbf{BAC-4B (ours)} &
        Chris &
        \textcolor{red}{$\times$} &

        \textbf{BAC-8B (ours)} &
        Chris &
        \textcolor{red}{$\times$} \\

        Qwen3-VL 2B &
        Chris &
        \textcolor{red}{$\times$} &

        Qwen3-VL 4B &
        Chris &
        \textcolor{red}{$\times$} &

        Qwen3-VL 8B &
        Chris &
        \textcolor{red}{$\times$} \\

        \midrule

        GPT-5.6 Sol &
        \texttt{UNID.} &
        \textcolor{green!60!black}{$\checkmark$} &

        GPT-5.6 Sol$^{*}$ &
        Chris &
        \textcolor{red}{$\times$} &

        Claude Sonnet 4.6 &
        \texttt{UNID.} &
        \textcolor{green!60!black}{$\checkmark$} \\

        Qwen3-VL 235B &
        Chris &
        \textcolor{red}{$\times$} &

        Gemini 2.5 Flash &
        Chris &
        \textcolor{red}{$\times$} &

        & & \\

        \bottomrule
    \end{tabular}

    \end{minipage}

    \vspace{1.2mm}

    \includegraphics[width=0.98\textwidth]
    {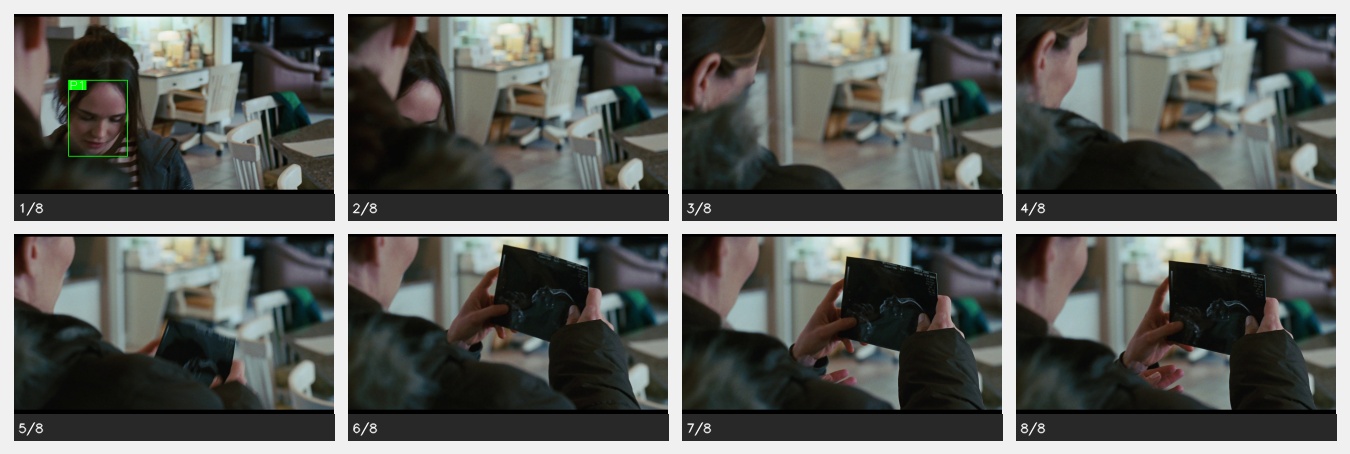}

    \vspace{0.3mm}

    \begin{minipage}{0.98\textwidth}
    \footnotesize

    \textbf{(c) Identity mapping:}
    \textcolor{green!60!black}{P1} $\leftrightarrow$
    \textcolor{green!60!black}{Elliot Page}.

    \vspace{0.2mm}

    \textbf{Q4 [Action]:}
    \textit{Who holds up an ultrasound image to look at it?}
    \hspace{0.6em}
    \textbf{GT:} \texttt{UNIDENTIFIED}.

    \vspace{0.3mm}

    \centering
    \scriptsize
    \setlength{\tabcolsep}{2.5pt}
    \renewcommand{\arraystretch}{0.90}

    \begin{tabular}{
        @{}lcc@{\hspace{8pt}}
        lcc@{\hspace{8pt}}
        lcc@{}
    }
        \toprule
        \textbf{Model} & \textbf{Pred.} & &
        \textbf{Model} & \textbf{Pred.} & &
        \textbf{Model} & \textbf{Pred.} & \\
        \midrule

        \textbf{BAC-2B (ours)} &
        \texttt{UNID.} &
        \textcolor{green!60!black}{$\checkmark$} &

        \textbf{BAC-4B (ours)} &
        \texttt{UNID.} &
        \textcolor{green!60!black}{$\checkmark$} &

        \textbf{BAC-8B (ours)} &
        \texttt{UNID.} &
        \textcolor{green!60!black}{$\checkmark$} \\

        Qwen3-VL 2B &
        Elliot &
        \textcolor{red}{$\times$} &

        Qwen3-VL 4B &
        \texttt{UNID.} &
        \textcolor{green!60!black}{$\checkmark$} &

        Qwen3-VL 8B &
        \texttt{UNID.} &
        \textcolor{green!60!black}{$\checkmark$} \\

        \midrule

        GPT-5.6 Sol &
        Elliot &
        \textcolor{red}{$\times$} &

        GPT-5.6 Sol$^{*}$ &
        Elliot &
        \textcolor{red}{$\times$} &

        Claude Sonnet 4.6 &
        \texttt{UNID.} &
        \textcolor{green!60!black}{$\checkmark$} \\

        Qwen3-VL 235B &
        \texttt{UNID.} &
        \textcolor{green!60!black}{$\checkmark$} &

        Gemini 2.5 Flash &
        \texttt{UNID.} &
        \textcolor{green!60!black}{$\checkmark$} &

        & & \\

        \bottomrule
    \end{tabular}

    \end{minipage}

    \vspace{-1.5mm}

    \captionof{figure}{Qualitative identity-aware QA examples (1/4).}
    \label{fig:qual_examples_1}

    \vspace{2mm}
}%

\clearpage

{%
\centering

    \includegraphics[width=0.98\textwidth]
    {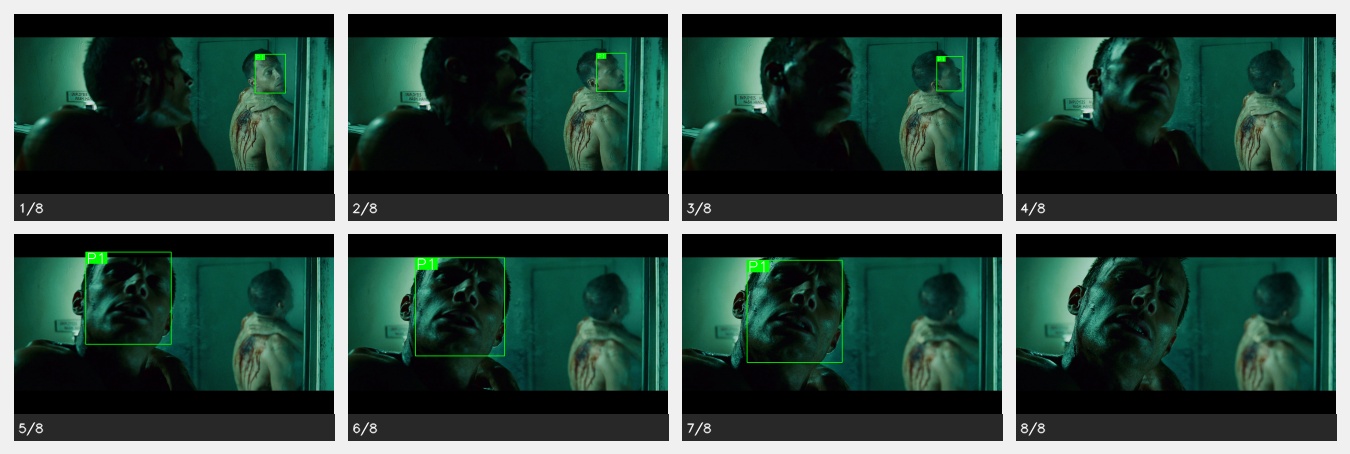}

    \vspace{0.3mm}

    \begin{minipage}{0.98\textwidth}
    \footnotesize

    \textbf{(a) Identity mapping:}
    \textcolor{green!60!black}{P1} $\leftrightarrow$
    \textcolor{green!60!black}{Paul Bettany}.

    \vspace{0.2mm}

    \textbf{Q1 [Appearance]:}
    \textit{Who is shirtless with blood dripping down his back?}
    \hspace{0.6em}
    \textbf{GT:}
    \textcolor{green!60!black}{Paul Bettany}
    (\textcolor{green!60!black}{P1}).

    \vspace{0.3mm}

    \centering
    \scriptsize
    \setlength{\tabcolsep}{2.5pt}
    \renewcommand{\arraystretch}{0.90}

    \begin{tabular}{
        @{}lcc@{\hspace{8pt}}
        lcc@{\hspace{8pt}}
        lcc@{}
    }
        \toprule
        \textbf{Model} & \textbf{Pred.} & &
        \textbf{Model} & \textbf{Pred.} & &
        \textbf{Model} & \textbf{Pred.} & \\
        \midrule

        \textbf{BAC-2B (ours)} &
        Paul &
        \textcolor{green!60!black}{$\checkmark$} &

        \textbf{BAC-4B (ours)} &
        Paul &
        \textcolor{green!60!black}{$\checkmark$} &

        \textbf{BAC-8B (ours)} &
        Paul &
        \textcolor{green!60!black}{$\checkmark$} \\

        Qwen3-VL 2B &
        Paul &
        \textcolor{green!60!black}{$\checkmark$} &

        Qwen3-VL 4B &
        Paul &
        \textcolor{green!60!black}{$\checkmark$} &

        Qwen3-VL 8B &
        Paul &
        \textcolor{green!60!black}{$\checkmark$} \\

        \midrule

        GPT-5.6 Sol &
        \texttt{UNID.} &
        \textcolor{red}{$\times$} &

        GPT-5.6 Sol$^{*}$ &
        Paul &
        \textcolor{green!60!black}{$\checkmark$} &

        Claude Sonnet 4.6 &
        \texttt{UNID.} &
        \textcolor{red}{$\times$} \\

        Qwen3-VL 235B &
        \texttt{UNID.} &
        \textcolor{red}{$\times$} &

        Gemini 2.5 Flash &
        \texttt{UNID.} &
        \textcolor{red}{$\times$} &

        & & \\

        \bottomrule
    \end{tabular}

    \end{minipage}

    \vspace{1.2mm}

    \includegraphics[width=0.98\textwidth]
    {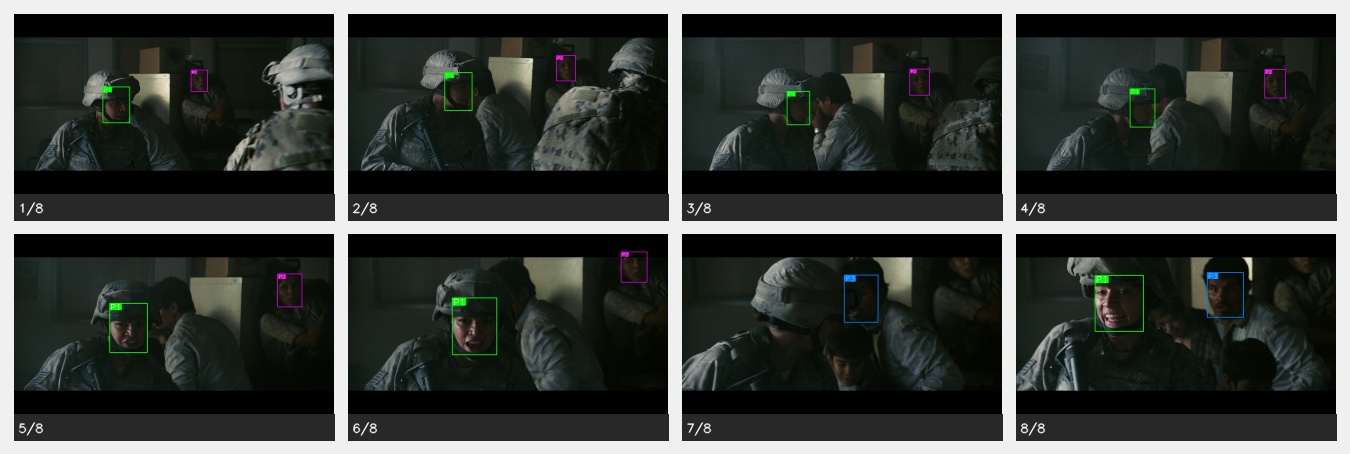}

    \vspace{0.3mm}

    \begin{minipage}{0.98\textwidth}
    \footnotesize

    \textbf{(b) Identity mapping:}
    \textcolor{green!60!black}{P1} $\leftrightarrow$
    \textcolor{green!60!black}{Michelle Rodriguez};
    \textcolor{magenta!80!black}{P2} $\leftrightarrow$
    \textcolor{magenta!80!black}{Bridget Moynahan};
    \textcolor{blue}{P3} $\leftrightarrow$
    \textcolor{blue}{Michael Pe\~{n}a}.

    \vspace{0.2mm}

    \textbf{Q1 [Posture]:}
    \textit{Who is leaning against the refrigerator or cabinet?}
    \hspace{0.6em}
    \textbf{GT:}
    \textcolor{magenta!80!black}{Bridget Moynahan}
    (\textcolor{magenta!80!black}{P2}).

    \vspace{0.3mm}

    \centering
    \scriptsize
    \setlength{\tabcolsep}{2.5pt}
    \renewcommand{\arraystretch}{0.90}

    \begin{tabular}{
        @{}lcc@{\hspace{8pt}}
        lcc@{\hspace{8pt}}
        lcc@{}
    }
        \toprule
        \textbf{Model} & \textbf{Pred.} & &
        \textbf{Model} & \textbf{Pred.} & &
        \textbf{Model} & \textbf{Pred.} & \\
        \midrule

        \textbf{BAC-2B (ours)} &
        Bridget &
        \textcolor{green!60!black}{$\checkmark$} &

        \textbf{BAC-4B (ours)} &
        Michael &
        \textcolor{red}{$\times$} &

        \textbf{BAC-8B (ours)} &
        Bridget &
        \textcolor{green!60!black}{$\checkmark$} \\

        Qwen3-VL 2B &
        Bridget &
        \textcolor{green!60!black}{$\checkmark$} &

        Qwen3-VL 4B &
        Bridget &
        \textcolor{green!60!black}{$\checkmark$} &

        Qwen3-VL 8B &
        Bridget &
        \textcolor{green!60!black}{$\checkmark$} \\

        \midrule

        GPT-5.6 Sol &
        Michael &
        \textcolor{red}{$\times$} &

        GPT-5.6 Sol$^{*}$ &
        Michael &
        \textcolor{red}{$\times$} &

        Claude Sonnet 4.6 &
        Bridget &
        \textcolor{green!60!black}{$\checkmark$} \\

        Qwen3-VL 235B &
        Bridget &
        \textcolor{green!60!black}{$\checkmark$} &

        Gemini 2.5 Flash &
        \texttt{UNID.} &
        \textcolor{red}{$\times$} &

        & & \\

        \bottomrule
    \end{tabular}

    \end{minipage}

    \vspace{1.2mm}

    \includegraphics[width=0.98\textwidth]
    {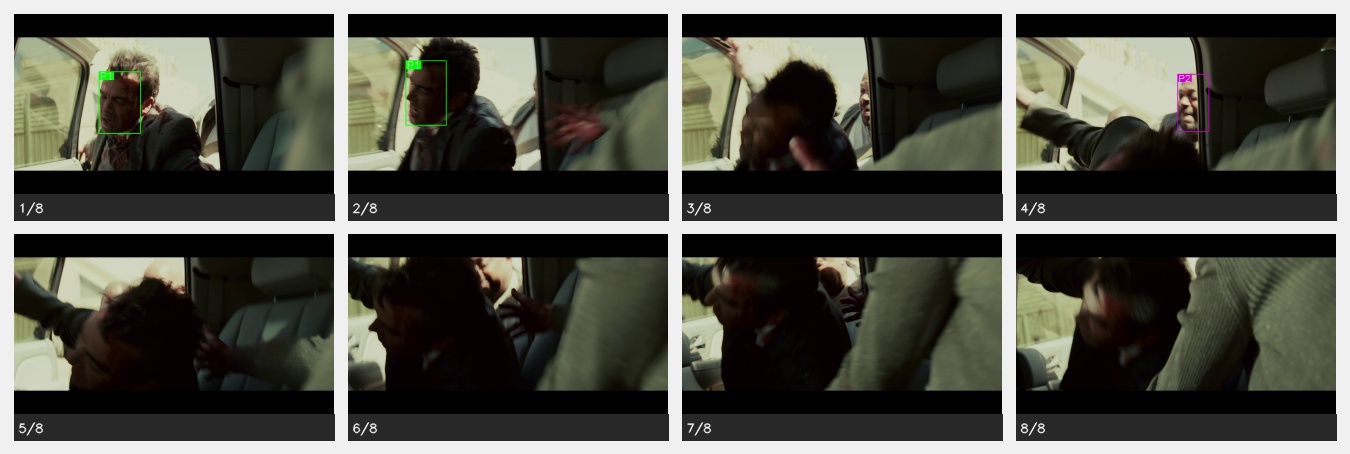}

    \vspace{0.3mm}

    \begin{minipage}{0.98\textwidth}
    \footnotesize

    \textbf{(c) Identity mapping:}
    \textcolor{green!60!black}{P1} $\leftrightarrow$
    \textcolor{green!60!black}{Jon Tenney};
    \textcolor{magenta!80!black}{P2} $\leftrightarrow$
    \textcolor{magenta!80!black}{Charles S. Dutton}.

    \vspace{0.2mm}

    \textbf{Q3 [Interaction]:}
    \textit{Who tries to pull Jon back into the vehicle?}
    \hspace{0.6em}
    \textbf{GT:}
    \textcolor{magenta!80!black}{Charles S. Dutton}
    (\textcolor{magenta!80!black}{P2}).

    \vspace{0.3mm}

    \centering
    \scriptsize
    \setlength{\tabcolsep}{2.5pt}
    \renewcommand{\arraystretch}{0.90}

    \begin{tabular}{
        @{}lcc@{\hspace{8pt}}
        lcc@{\hspace{8pt}}
        lcc@{}
    }
        \toprule
        \textbf{Model} & \textbf{Pred.} & &
        \textbf{Model} & \textbf{Pred.} & &
        \textbf{Model} & \textbf{Pred.} & \\
        \midrule

        \textbf{BAC-2B (ours)} &
        Charles &
        \textcolor{green!60!black}{$\checkmark$} &

        \textbf{BAC-4B (ours)} &
        Charles &
        \textcolor{green!60!black}{$\checkmark$} &

        \textbf{BAC-8B (ours)} &
        Charles &
        \textcolor{green!60!black}{$\checkmark$} \\

        Qwen3-VL 2B &
        Charles &
        \textcolor{green!60!black}{$\checkmark$} &

        Qwen3-VL 4B &
        \texttt{UNID.} &
        \textcolor{red}{$\times$} &

        Qwen3-VL 8B &
        Charles &
        \textcolor{green!60!black}{$\checkmark$} \\

        \midrule

        GPT-5.6 Sol &
        Charles &
        \textcolor{green!60!black}{$\checkmark$} &

        GPT-5.6 Sol$^{*}$ &
        \texttt{UNID.} &
        \textcolor{red}{$\times$} &

        Claude Sonnet 4.6 &
        \texttt{UNID.} &
        \textcolor{red}{$\times$} \\

        Qwen3-VL 235B &
        \texttt{UNID.} &
        \textcolor{red}{$\times$} &

        Gemini 2.5 Flash &
        \texttt{UNID.} &
        \textcolor{red}{$\times$} &

        & & \\

        \bottomrule
    \end{tabular}

    \end{minipage}

    \vspace{-1.5mm}

    \captionof{figure}{Qualitative identity-aware QA examples (2/4).}
    \label{fig:qual_examples_2}

    \vspace{2mm}
}%

\clearpage

{%
\centering

    \includegraphics[width=0.98\textwidth]
    {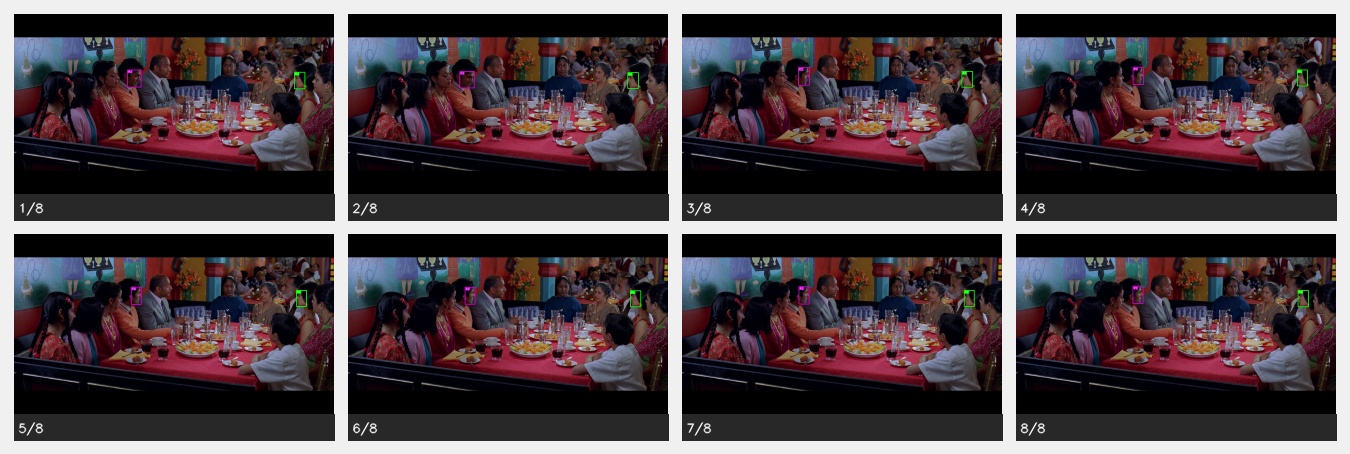}

    \vspace{0.3mm}

    \begin{minipage}{0.98\textwidth}
    \footnotesize

    \textbf{(a) Identity mapping:}
    \textcolor{green!60!black}{P1} $\leftrightarrow$
    \textcolor{green!60!black}{Anjali Jay};
    \textcolor{magenta!80!black}{P2} $\leftrightarrow$
    \textcolor{magenta!80!black}{Sendhil Ramamurthy}.

    \vspace{0.2mm}

    \textbf{Q2 [Appearance]:}
    \textit{Who is wearing a red sleeveless dress?}
    \hspace{0.6em}
    \textbf{GT:}
    \textcolor{green!60!black}{Anjali Jay}
    (\textcolor{green!60!black}{P1}).

    \vspace{0.3mm}

    \centering
    \scriptsize
    \setlength{\tabcolsep}{2.5pt}
    \renewcommand{\arraystretch}{0.90}

    \begin{tabular}{
        @{}lcc@{\hspace{8pt}}
        lcc@{\hspace{8pt}}
        lcc@{}
    }
        \toprule
        \textbf{Model} & \textbf{Pred.} & &
        \textbf{Model} & \textbf{Pred.} & &
        \textbf{Model} & \textbf{Pred.} & \\
        \midrule

        \textbf{BAC-2B (ours)} &
        Anjali &
        \textcolor{green!60!black}{$\checkmark$} &

        \textbf{BAC-4B (ours)} &
        Anjali &
        \textcolor{green!60!black}{$\checkmark$} &

        \textbf{BAC-8B (ours)} &
        \texttt{UNID.} &
        \textcolor{red}{$\times$} \\

        Qwen3-VL 2B &
        Anjali &
        \textcolor{green!60!black}{$\checkmark$} &

        Qwen3-VL 4B &
        \texttt{UNID.} &
        \textcolor{red}{$\times$} &

        Qwen3-VL 8B &
        \texttt{UNID.} &
        \textcolor{red}{$\times$} \\

        \midrule

        GPT-5.6 Sol &
        \texttt{UNID.} &
        \textcolor{red}{$\times$} &

        GPT-5.6 Sol$^{*}$ &
        Anjali &
        \textcolor{green!60!black}{$\checkmark$} &

        Claude Sonnet 4.6 &
        \texttt{UNID.} &
        \textcolor{red}{$\times$} \\

        Qwen3-VL 235B &
        \texttt{UNID.} &
        \textcolor{red}{$\times$} &

        Gemini 2.5 Flash &
        \texttt{UNID.} &
        \textcolor{red}{$\times$} &

        & & \\

        \bottomrule
    \end{tabular}

    \end{minipage}

    \vspace{1.2mm}

    \includegraphics[width=0.98\textwidth]
    {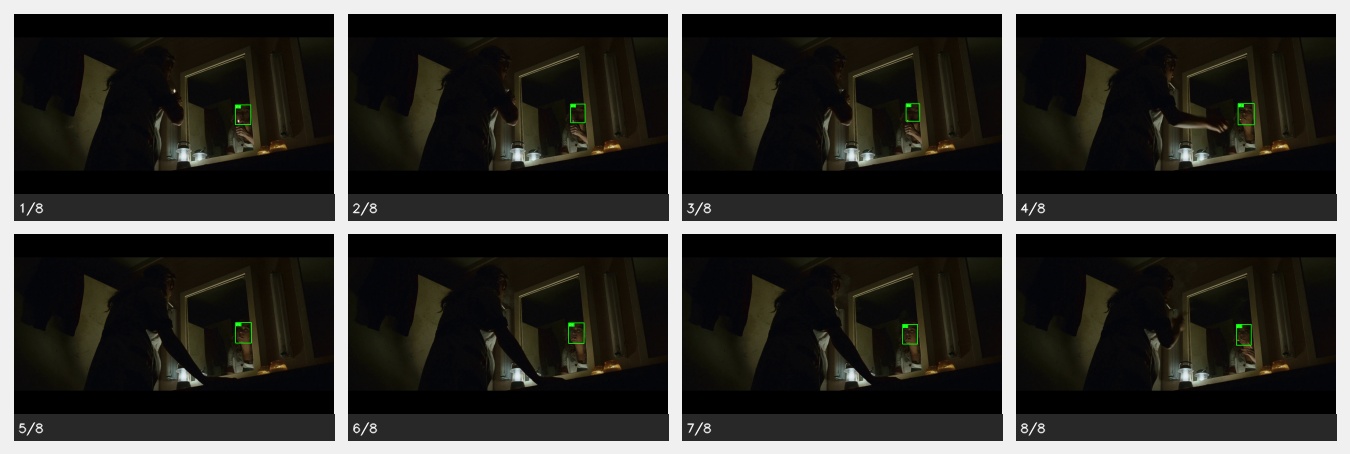}

    \vspace{0.3mm}

    \begin{minipage}{0.98\textwidth}
    \footnotesize

    \textbf{(b) Identity mapping:}
    \textcolor{green!60!black}{P1} $\leftrightarrow$
    \textcolor{green!60!black}{Adrianne Palicki}.

    \vspace{0.2mm}

    \textbf{Q3 [Posture]:}
    \textit{Who is leaning against the sink?}
    \hspace{0.6em}
    \textbf{GT:}
    \textcolor{green!60!black}{Adrianne Palicki}
    (\textcolor{green!60!black}{P1}).

    \vspace{0.3mm}

    \centering
    \scriptsize
    \setlength{\tabcolsep}{2.5pt}
    \renewcommand{\arraystretch}{0.90}

    \begin{tabular}{
        @{}lcc@{\hspace{8pt}}
        lcc@{\hspace{8pt}}
        lcc@{}
    }
        \toprule
        \textbf{Model} & \textbf{Pred.} & &
        \textbf{Model} & \textbf{Pred.} & &
        \textbf{Model} & \textbf{Pred.} & \\
        \midrule

        \textbf{BAC-2B (ours)} &
        Adrianne &
        \textcolor{green!60!black}{$\checkmark$} &

        \textbf{BAC-4B (ours)} &
        Adrianne &
        \textcolor{green!60!black}{$\checkmark$} &

        \textbf{BAC-8B (ours)} &
        Adrianne &
        \textcolor{green!60!black}{$\checkmark$} \\

        Qwen3-VL 2B &
        Adrianne &
        \textcolor{green!60!black}{$\checkmark$} &

        Qwen3-VL 4B &
        Adrianne &
        \textcolor{green!60!black}{$\checkmark$} &

        Qwen3-VL 8B &
        Adrianne &
        \textcolor{green!60!black}{$\checkmark$} \\

        \midrule

        GPT-5.6 Sol &
        \texttt{UNID.} &
        \textcolor{red}{$\times$} &

        GPT-5.6 Sol$^{*}$ &
        Adrianne &
        \textcolor{green!60!black}{$\checkmark$} &

        Claude Sonnet 4.6 &
        \texttt{UNID.} &
        \textcolor{red}{$\times$} \\

        Qwen3-VL 235B &
        Adrianne &
        \textcolor{green!60!black}{$\checkmark$} &

        Gemini 2.5 Flash &
        \texttt{UNID.} &
        \textcolor{red}{$\times$} &

        & & \\

        \bottomrule
    \end{tabular}

    \end{minipage}

    \vspace{1.2mm}

    \includegraphics[width=0.98\textwidth]
    {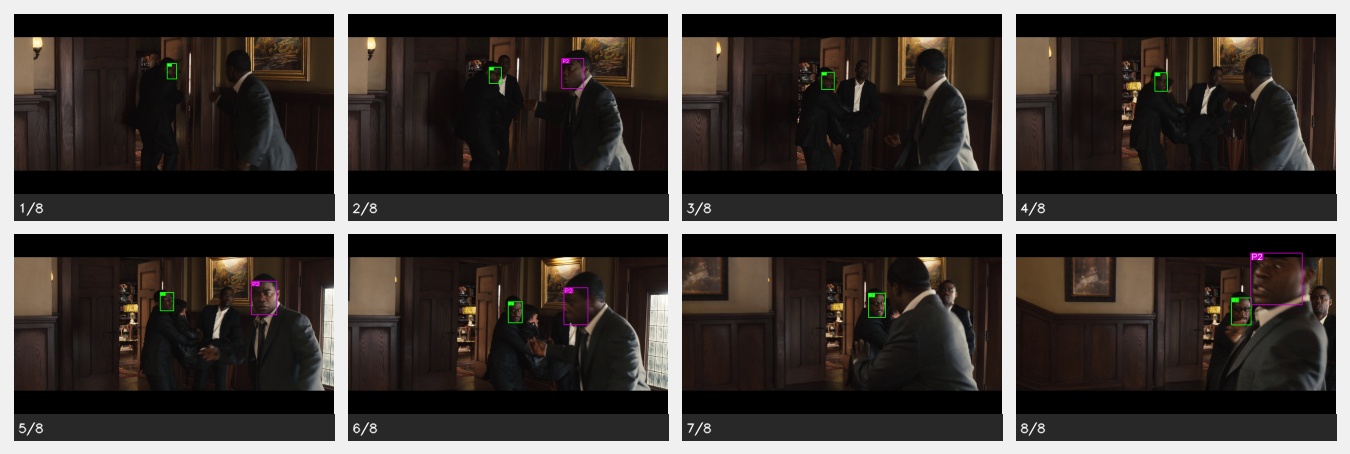}

    \vspace{0.3mm}

    \begin{minipage}{0.98\textwidth}
    \footnotesize

    \textbf{(c) Identity mapping:}
    \textcolor{green!60!black}{P1} $\leftrightarrow$
    \textcolor{green!60!black}{Martin Lawrence};
    \textcolor{magenta!80!black}{P2} $\leftrightarrow$
    \textcolor{magenta!80!black}{Tracy Morgan}.

    \vspace{0.2mm}

    \textbf{Q4 [Interaction]:}
    \textit{Who is holding a young man from behind, supporting him under the arms?}
    \hspace{0.6em}
    \textbf{GT:}
    \textcolor{green!60!black}{Martin Lawrence}
    (\textcolor{green!60!black}{P1}).

    \vspace{0.3mm}

    \centering
    \scriptsize
    \setlength{\tabcolsep}{2.5pt}
    \renewcommand{\arraystretch}{0.90}

    \begin{tabular}{
        @{}lcc@{\hspace{8pt}}
        lcc@{\hspace{8pt}}
        lcc@{}
    }
        \toprule
        \textbf{Model} & \textbf{Pred.} & &
        \textbf{Model} & \textbf{Pred.} & &
        \textbf{Model} & \textbf{Pred.} & \\
        \midrule

        \textbf{BAC-2B (ours)} &
        Martin &
        \textcolor{green!60!black}{$\checkmark$} &

        \textbf{BAC-4B (ours)} &
        Martin &
        \textcolor{green!60!black}{$\checkmark$} &

        \textbf{BAC-8B (ours)} &
        Martin &
        \textcolor{green!60!black}{$\checkmark$} \\

        Qwen3-VL 2B &
        Martin &
        \textcolor{green!60!black}{$\checkmark$} &

        Qwen3-VL 4B &
        Martin &
        \textcolor{green!60!black}{$\checkmark$} &

        Qwen3-VL 8B &
        Martin &
        \textcolor{green!60!black}{$\checkmark$} \\

        \midrule

        GPT-5.6 Sol &
        Martin &
        \textcolor{green!60!black}{$\checkmark$} &

        GPT-5.6 Sol$^{*}$ &
        \texttt{UNID.} &
        \textcolor{red}{$\times$} &

        Claude Sonnet 4.6 &
        \texttt{UNID.} &
        \textcolor{red}{$\times$} \\

        Qwen3-VL 235B &
        \texttt{UNID.} &
        \textcolor{red}{$\times$} &

        Gemini 2.5 Flash &
        Martin &
        \textcolor{green!60!black}{$\checkmark$} &

        & & \\

        \bottomrule
    \end{tabular}

    \end{minipage}

    \vspace{-1.5mm}

    \captionof{figure}{Qualitative identity-aware QA examples (3/4).}
    \label{fig:qual_examples_3}

    \vspace{2mm}
}%

\clearpage

{%
\centering

    \includegraphics[width=0.98\textwidth]
    {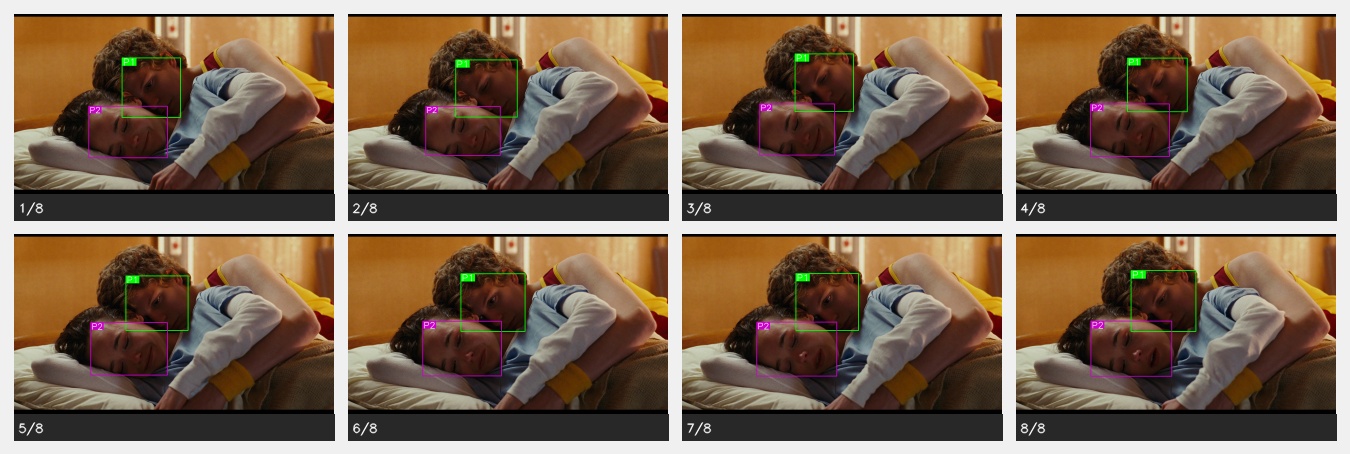}

    \vspace{0.3mm}

    \begin{minipage}{0.98\textwidth}
    \footnotesize

    \textbf{(a) Identity mapping:}
    \textcolor{green!60!black}{P1} $\leftrightarrow$
    \textcolor{green!60!black}{Michael Cera};
    \textcolor{magenta!80!black}{P2} $\leftrightarrow$
    \textcolor{magenta!80!black}{Elliot Page}.

    \vspace{0.2mm}

    \textbf{Q2 [Appearance]:}
    \textit{Who is wearing a light blue shirt?}
    \hspace{0.6em}
    \textbf{GT:}
    \textcolor{magenta!80!black}{Elliot Page}
    (\textcolor{magenta!80!black}{P2}).

    \textbf{Q4 [Posture]:}
    \textit{Who rests against a pillow with closed eyes before opening them?}
    \hspace{0.6em}
    \textbf{GT:}
    \textcolor{magenta!80!black}{Elliot Page}
    (\textcolor{magenta!80!black}{P2}).

    \vspace{0.3mm}

    \centering
    \scriptsize
    \setlength{\tabcolsep}{2.5pt}
    \renewcommand{\arraystretch}{0.90}

    \begin{tabular}{
        @{}lccc@{\hspace{6pt}}
        lccc@{\hspace{6pt}}
        lccc@{}
    }
        \toprule
        \textbf{Model} & \textbf{Q2} & \textbf{Q4} & &
        \textbf{Model} & \textbf{Q2} & \textbf{Q4} & &
        \textbf{Model} & \textbf{Q2} & \textbf{Q4} & \\
        \midrule

        \textbf{BAC-2B (ours)} &
        \textcolor{green!60!black}{$\checkmark$} &
        \textcolor{green!60!black}{$\checkmark$} & &

        \textbf{BAC-4B (ours)} &
        \textcolor{green!60!black}{$\checkmark$} &
        \textcolor{green!60!black}{$\checkmark$} & &

        \textbf{BAC-8B (ours)} &
        \textcolor{green!60!black}{$\checkmark$} &
        \textcolor{green!60!black}{$\checkmark$} & \\

        Qwen3-VL 2B &
        Michael \textcolor{red}{$\times$} &
        \textcolor{green!60!black}{$\checkmark$} & &

        Qwen3-VL 4B &
        \textcolor{green!60!black}{$\checkmark$} &
        \textcolor{green!60!black}{$\checkmark$} & &

        Qwen3-VL 8B &
        \textcolor{green!60!black}{$\checkmark$} &
        \textcolor{green!60!black}{$\checkmark$} & \\

        \midrule

        GPT-5.6 Sol &
        Michael \textcolor{red}{$\times$} &
        Michael \textcolor{red}{$\times$} & &

        GPT-5.6 Sol$^{*}$ &
        \textcolor{green!60!black}{$\checkmark$} &
        \textcolor{green!60!black}{$\checkmark$} & &

        Claude Sonnet 4.6 &
        \textcolor{green!60!black}{$\checkmark$} &
        Michael \textcolor{red}{$\times$} & \\

        Qwen3-VL 235B &
        \textcolor{green!60!black}{$\checkmark$} &
        \textcolor{green!60!black}{$\checkmark$} & &

        Gemini 2.5 Flash &
        \texttt{UNID.} \textcolor{red}{$\times$} &
        \textcolor{green!60!black}{$\checkmark$} & &

        & & & \\

        \bottomrule
    \end{tabular}

    \end{minipage}

    \vspace{1.2mm}

    \includegraphics[width=0.98\textwidth]
    {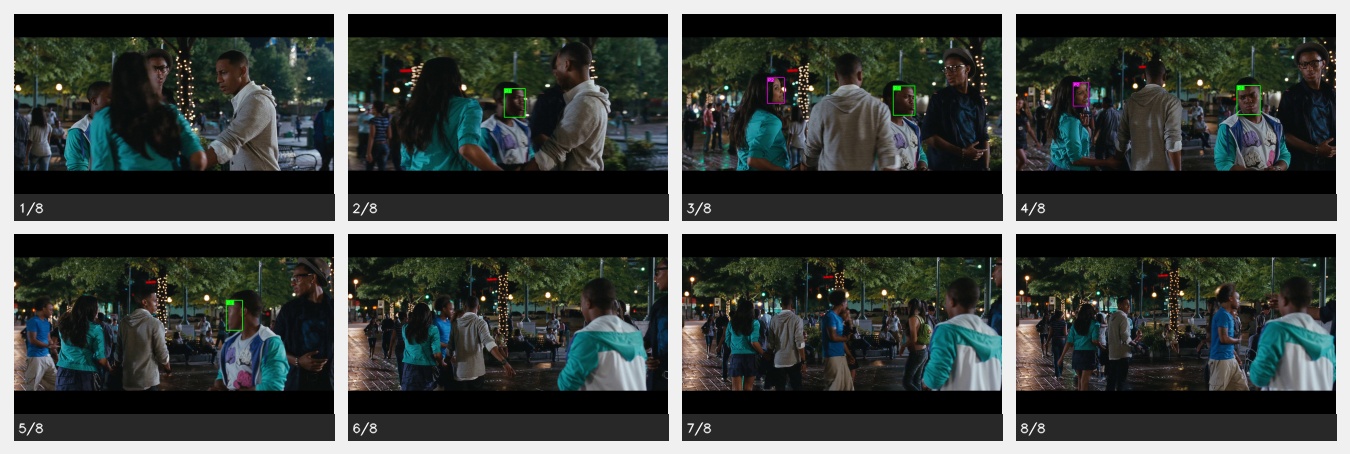}

    \vspace{0.3mm}

    \begin{minipage}{0.98\textwidth}
    \footnotesize

    \textbf{(b) Identity mapping:}
    \textcolor{green!60!black}{P1} $\leftrightarrow$
    \textcolor{green!60!black}{Marc John Jefferies};
    \textcolor{magenta!80!black}{P2} $\leftrightarrow$
    \textcolor{magenta!80!black}{Jessica Lucas}.

    \vspace{0.2mm}

    \textbf{Q4 [Interaction]:}
    \textit{Who turns back to look while walking away hand-in-hand with the man in the gray hoodie?}
    \hspace{0.6em}
    \textbf{GT:}
    \textcolor{magenta!80!black}{Jessica Lucas}
    (\textcolor{magenta!80!black}{P2}).

    \vspace{0.3mm}

    \centering
    \scriptsize
    \setlength{\tabcolsep}{2.5pt}
    \renewcommand{\arraystretch}{0.90}

    \begin{tabular}{
        @{}lcc@{\hspace{8pt}}
        lcc@{\hspace{8pt}}
        lcc@{}
    }
        \toprule
        \textbf{Model} & \textbf{Pred.} & &
        \textbf{Model} & \textbf{Pred.} & &
        \textbf{Model} & \textbf{Pred.} & \\
        \midrule

        \textbf{BAC-2B (ours)} &
        Jessica &
        \textcolor{green!60!black}{$\checkmark$} &

        \textbf{BAC-4B (ours)} &
        Jessica &
        \textcolor{green!60!black}{$\checkmark$} &

        \textbf{BAC-8B (ours)} &
        Jessica &
        \textcolor{green!60!black}{$\checkmark$} \\

        Qwen3-VL 2B &
        Marc &
        \textcolor{red}{$\times$} &

        Qwen3-VL 4B &
        Jessica &
        \textcolor{green!60!black}{$\checkmark$} &

        Qwen3-VL 8B &
        Jessica &
        \textcolor{green!60!black}{$\checkmark$} \\

        \midrule

        GPT-5.6 Sol &
        \texttt{UNID.} &
        \textcolor{red}{$\times$} &

        GPT-5.6 Sol$^{*}$ &
        Jessica &
        \textcolor{green!60!black}{$\checkmark$} &

        Claude Sonnet 4.6 &
        Jessica &
        \textcolor{green!60!black}{$\checkmark$} \\

        Qwen3-VL 235B &
        Jessica &
        \textcolor{green!60!black}{$\checkmark$} &

        Gemini 2.5 Flash &
        \texttt{UNID.} &
        \textcolor{red}{$\times$} &

        & & \\

        \bottomrule
    \end{tabular}

    \end{minipage}

    \vspace{1.2mm}

    \includegraphics[width=0.98\textwidth]
    {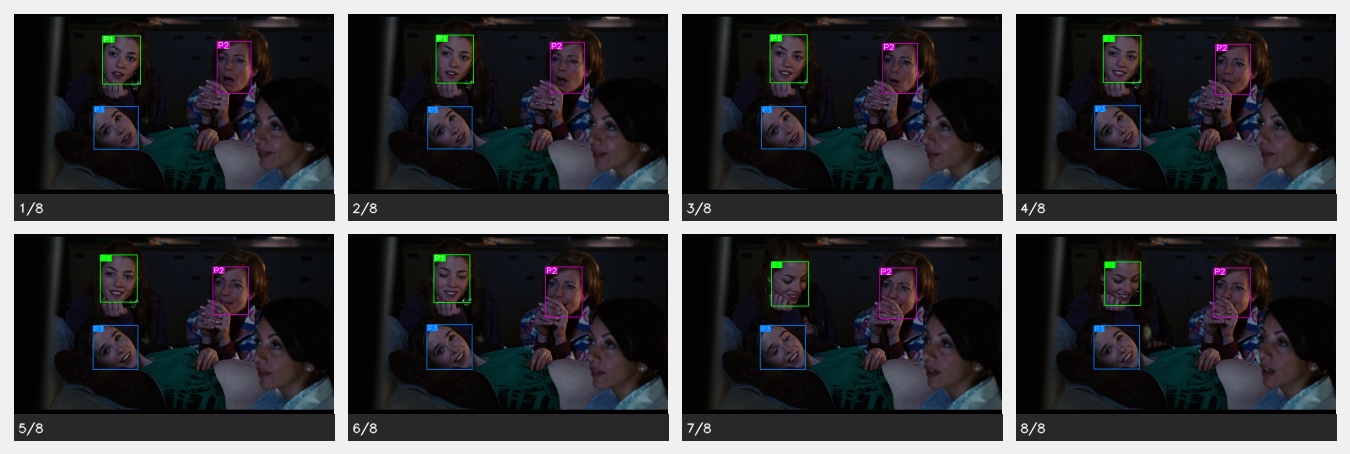}

    \vspace{0.3mm}

    \begin{minipage}{0.98\textwidth}
    \footnotesize

    \textbf{(c) Identity mapping:}
    \textcolor{green!60!black}{P1} $\leftrightarrow$
    \textcolor{green!60!black}{Olivia Thirlby};
    \textcolor{magenta!80!black}{P2} $\leftrightarrow$
    \textcolor{magenta!80!black}{Allison Janney};
    \textcolor{blue}{P3} $\leftrightarrow$
    \textcolor{blue}{Elliot Page}.

    \vspace{0.2mm}

    \textbf{Q1 [Appearance]:}
    \textit{Who is reclining with an exposed pregnant belly?}
    \hspace{0.6em}
    \textbf{GT:}
    \textcolor{blue}{Elliot Page}
    (\textcolor{blue}{P3}).

    \vspace{0.3mm}

    \centering
    \scriptsize
    \setlength{\tabcolsep}{2.5pt}
    \renewcommand{\arraystretch}{0.90}

    \begin{tabular}{
        @{}lcc@{\hspace{8pt}}
        lcc@{\hspace{8pt}}
        lcc@{}
    }
        \toprule
        \textbf{Model} & \textbf{Pred.} & &
        \textbf{Model} & \textbf{Pred.} & &
        \textbf{Model} & \textbf{Pred.} & \\
        \midrule

        \textbf{BAC-2B (ours)} &
        Elliot &
        \textcolor{green!60!black}{$\checkmark$} &

        \textbf{BAC-4B (ours)} &
        Elliot &
        \textcolor{green!60!black}{$\checkmark$} &

        \textbf{BAC-8B (ours)} &
        Allison &
        \textcolor{red}{$\times$} \\

        Qwen3-VL 2B &
        Elliot &
        \textcolor{green!60!black}{$\checkmark$} &

        Qwen3-VL 4B &
        Allison &
        \textcolor{red}{$\times$} &

        Qwen3-VL 8B &
        Allison &
        \textcolor{red}{$\times$} \\

        \midrule

        GPT-5.6 Sol &
        \texttt{UNID.} &
        \textcolor{red}{$\times$} &

        GPT-5.6 Sol$^{*}$ &
        Elliot &
        \textcolor{green!60!black}{$\checkmark$} &

        Claude Sonnet 4.6 &
        \texttt{UNID.} &
        \textcolor{red}{$\times$} \\

        Qwen3-VL 235B &
        Elliot &
        \textcolor{green!60!black}{$\checkmark$} &

        Gemini 2.5 Flash &
        Elliot &
        \textcolor{green!60!black}{$\checkmark$} &

        & & \\

        \bottomrule
    \end{tabular}

    \end{minipage}

    \vspace{-1.5mm}

    \captionof{figure}{Qualitative identity-aware QA examples (4/4).}
    \label{fig:qual_examples_4}

    \vspace{2mm}
}%

%% file: iclr2027_conference.bib
@article{lsmdc2016MovieAnnotationRetrieval, 
author = {Torabi, Atousa and Tandon, Niket and Sigal,Leon}, 
title = {Learning Language-Visual Embedding for Movie Understanding with Natural-Language}, 
journal = {arXiv:1609.08124},
year = {2016}, 
url = {http://arxiv.org/pdf/1609.08124v1.pdf}}

@inproceedings{maharaj2017dataset,
  title={A dataset and exploration of models for understanding video data through fill-in-the-blank question-answering.},
  author={Maharaj, Tegan and Ballas, Nicolas and Rohrbach, Anna and Courville, Aaron C and Pal, Christopher Joseph},
  booktitle={Computer Vision and Pattern Recognition (CVPR)},
  year={2017},
  url={http://openaccess.thecvf.com/content_cvpr_2017/papers/Maharaj_A_Dataset_and_CVPR_2017_paper.pdf}}

@article{lsmdc,
author = {Rohrbach, Anna and Torabi, Atousa and Rohrbach, Marcus and Tandon, Niket and  Pal, Chris and Larochelle, Hugo and Courville, Aaron and Schiele, Bernt},
title = {Movie Description},
journal={International Journal of Computer Vision},
year = {2017},
url = {http://link.springer.com/article/10.1007/s11263-016-0987-1?wt_mc=Internal.Event.1.SEM.ArticleAuthorOnlineFirst}}

@inproceedings{xu2010cast2face,
  title={Cast2face: Character identification in movie with actor-character correspondence},
  author={Xu, Mengdi and Yuan, Xiaotong and Shen, Jialie and Yan, Shuicheng},
  booktitle={Proceedings of the 18th ACM international conference on Multimedia},
  pages={831--834},
  year={2010}
}

@article{nagrani2018benedict,
  title={From benedict cumberbatch to sherlock holmes: Character identification in tv series without a script},
  author={Nagrani, Arsha and Zisserman, Andrew},
  journal={arXiv preprint arXiv:1801.10442},
  year={2018}
}

@inproceedings{raajesh2024micap,
  title={Micap: A unified model for identity-aware movie descriptions},
  author={Raajesh, Haran and Desanur, Naveen Reddy and Khan, Zeeshan and Tapaswi, Makarand},
  booktitle={2024 IEEE/CVF Conference on Computer Vision and Pattern Recognition (CVPR)},
  pages={14011--14021},
  year={2024},
  organization={IEEE}
}

@article{bamman2024measuring,
  title={Measuring diversity in Hollywood through the large-scale computational analysis of film},
  author={Bamman, David and Samberg, Rachael and So, Richard Jean and Zhou, Naitian},
  journal={Proceedings of the National Academy of Sciences},
  volume={121},
  number={46},
  pages={e2409770121},
  year={2024},
  publisher={National Academy of Sciences}
}

@inproceedings{ji2025ida,
  title={IDA-VLM: towards movie understanding via id-aware large vision-language model},
  author={Ji, Yatai and Zhang, Shilong and Wu, Jie and Sun, Peize and Chen, Weifeng and Xiao, Xuefeng and Yang, Sidi and Yang, Yujiu and Luo, Ping},
  booktitle={International Conference on Learning Representations},
  volume={2025},
  pages={52639--52652},
  year={2025}
}

@article{gao2026seek,
  title={I Seek You in Videos: Identity-Conditioned Queries for Person-Centric Video Reasoning},
  author={Gao, Shibo and Wang, Chongxiao and Huang, Chenglong and Ma, Jie and Shi, Haolin and Ding, Fei and Li, Jing and Lyu, Qiang and Liu, Yangyang and Liu, Yang and others},
  journal={arXiv preprint arXiv:2608.07417},
  year={2026}
}

@inproceedings{heo2025omni,
  title={Omni-rgpt: Unifying image and video region-level understanding via token marks},
  author={Heo, Miran and Chen, Min-Hung and Huang, De-An and Liu, Sifei and Radhakrishnan, Subhashree and Kim, Seon Joo and Wang, Yu-Chiang Frank and Hachiuma, Ryo},
  booktitle={2025 IEEE/CVF Conference on Computer Vision and Pattern Recognition (CVPR)},
  pages={3919--3930},
  year={2025},
  organization={IEEE}
}

@article{pini2019m,
  title={M-VAD names: a dataset for video captioning with naming},
  author={Pini, Stefano and Cornia, Marcella and Bolelli, Federico and Baraldi, Lorenzo and Cucchiara, Rita},
  journal={Multimedia Tools and Applications},
  volume={78},
  number={10},
  pages={14007--14027},
  year={2019},
  publisher={Springer}
}

@inproceedings{ma2024groma,
  title={Groma: Localized visual tokenization for grounding multimodal large language models},
  author={Ma, Chuofan and Jiang, Yi and Wu, Jiannan and Yuan, Zehuan and Qi, Xiaojuan},
  booktitle={European Conference on Computer Vision},
  pages={417--435},
  year={2024},
  organization={Springer}
}

@InProceedings{Kamath_2021_ICCV,
    author    = {Kamath, Aishwarya and Singh, Mannat and LeCun, Yann and Synnaeve, Gabriel and Misra, Ishan and Carion, Nicolas},
    title     = {MDETR - Modulated Detection for End-to-End Multi-Modal Understanding},
    booktitle = {Proceedings of the IEEE/CVF International Conference on Computer Vision (ICCV)},
    month     = {October},
    year      = {2021},
    pages     = {1780-1790}
}

@InProceedings{Yang_2023_CVPR,
    author    = {Yang, Antoine and Nagrani, Arsha and Seo, Paul Hongsuck and Miech, Antoine and Pont-Tuset, Jordi and Laptev, Ivan and Sivic, Josef and Schmid, Cordelia},
    title     = {Vid2Seq: Large-Scale Pretraining of a Visual Language Model for Dense Video Captioning},
    booktitle = {Proceedings of the IEEE/CVF Conference on Computer Vision and Pattern Recognition (CVPR)},
    month     = {June},
    year      = {2023},
    pages     = {10714-10726}
}

@inproceedings{xu2016msr,
  title={Msr-vtt: A large video description dataset for bridging video and language},
  author={Xu, Jun and Mei, Tao and Yao, Ting and Rui, Yong},
  booktitle={Proceedings of the IEEE conference on computer vision and pattern recognition},
  pages={5288--5296},
  year={2016}
}

@article{jaruenpunyasak2022anthropometric,
  title={Anthropometric ratios for lower-body detection based on deep learning and traditional methods},
  author={Jaruenpunyasak, Jermphiphut and Garc{\'\i}a Seco de Herrera, Alba and Duangsoithong, Rakkrit},
  journal={Applied Sciences},
  volume={12},
  number={5},
  pages={2678},
  year={2022},
  publisher={MDPI}
}

@article{tevissen2024inserting,
  title={Inserting faces inside captions: image captioning with attention guided merging},
  author={Tevissen, Yannis and Guetari, Khalil and Tassel, Marine and Kerleroux, Erwan and Petitpont, Fr{\'e}d{\'e}ric},
  journal={arXiv preprint arXiv:2405.02305},
  year={2024}
}

@inproceedings{munasinghe2025videoglamm,
  title={Videoglamm: A large multimodal model for pixel-level visual grounding in videos},
  author={Munasinghe, Shehan and Gani, Hanan and Zhu, Wenqi and Cao, Jiale and Xing, Eric and Khan, Fahad Shahbaz and Khan, Salman},
  booktitle={2025 IEEE/CVF Conference on Computer Vision and Pattern Recognition (CVPR)},
  pages={19036--19046},
  year={2025},
  organization={IEEE}
}

@inproceedings{deng2019arcface,
  title={Arcface: Additive angular margin loss for deep face recognition},
  author={Deng, Jiankang and Guo, Jia and Xue, Niannan and Zafeiriou, Stefanos},
  booktitle={Proceedings of the IEEE/CVF conference on computer vision and pattern recognition},
  pages={4690--4699},
  year={2019}
}

@inproceedings{wojke2017simple,
  title={Simple online and realtime tracking with a deep association metric},
  author={Wojke, Nicolai and Bewley, Alex and Paulus, Dietrich},
  booktitle={2017 IEEE international conference on image processing (ICIP)},
  pages={3645--3649},
  year={2017},
  organization={IEEE}
}
